\documentclass[final,5p,times,twocolumn]{elsarticle}

\usepackage{pgfplots}
\usepackage{newfloat}
\usepackage{listings}
\usepackage{algorithm}
\usepackage{alltt}
\usepackage{upquote}
\usepackage{booktabs}

\usepackage{amssymb,amsmath}
\usepackage{bm}
\usepackage{color}
\usepackage{dcolumn}
\usepackage{graphicx}
\usepackage{multirow}
\usepackage{subfigure}
\usepackage{kotex}
\usepackage{enumitem}
\usepackage{lineno}
\usepackage{adjustbox}

\journal{Knowledge-Based Systems}
\date{}

\begin{document}

\begin{frontmatter}



\title{Maximizing Discrimination Capability of Knowledge Distillation with Energy Function}

\author[1]{Seonghak Kim\corref{cont1}}
\ead{hakk35@kaist.ac.kr}

\author[1]{Gyeongdo Ham\corref{cont1}}
\ead{rudeh6185@kaist.ac.kr}

\author[1]{Suin Lee}
\ead{suinlee@kaist.ac.kr}

\author[1]{Donggon Jang}
\ead{jdg900@kaist.ac.kr}

\author[1]{Daeshik Kim\corref{cores}}
\ead{daeshik@kaist.ac.kr}

\affiliation[1]{organization={School of Electrical Engineering, Korea Advanced Institute of Science and Technology},
            addressline={291, Daehak-ro, Yuseong-gu}, 
            city={Daejeon},
            postcode={34141}, 
            country={Republic of Korea}}

\cortext[cont1]{Authors contributed equally}
\cortext[cores]{Corresponding author}

\tnotetext[]{\copyright~2024 This is the author’s accepted manuscript of an article published in Knowledge-Based Systems, made available under the CC-BY-NC-ND 4.0 license: https://creativecommons.org/licenses/by-nc-nd/4.0/ \\ Digital Object Identifier (DOI): 10.1016/j.knosys.2024.111911}

\begin{abstract}
To apply the latest computer vision techniques that require a large computational cost in real industrial applications, knowledge distillation methods (KDs) are essential. Existing logit-based KDs apply the constant temperature scaling to all samples in dataset, limiting the utilization of knowledge inherent in each sample individually. In our approach, we classify the dataset into two categories (i.e., low energy and high energy samples) based on their energy score. Through experiments, we have confirmed that low energy samples exhibit high confidence scores, indicating certain predictions, while high energy samples yield low confidence scores, meaning uncertain predictions. To distill optimal knowledge by adjusting non-target class predictions, we apply a higher temperature to low energy samples to create smoother distributions and a lower temperature to high energy samples to achieve sharper distributions. When compared to previous logit-based and feature-based methods, our energy-based KD (Energy KD) achieves better performance on various datasets. Especially, Energy KD shows significant improvements on CIFAR-100-LT and ImageNet datasets, which contain many challenging samples. Furthermore, we propose high energy-based data augmentation (HE-DA) for further improving the performance. We demonstrate that higher performance improvement could be achieved by augmenting only a portion of the dataset rather than the entire dataset, suggesting that it can be employed on resource-limited devices. To the best of our knowledge, this paper represents the first attempt to make use of energy function in knowledge distillation and data augmentation, and we believe it will greatly contribute to future research.
\end{abstract}




\end{frontmatter}


\section{Introduction}
\label{sec:introduction}
In recent years, computer vision has witnessed significant advancements, notably in areas like image classification~\cite{class1, class2}, object detection~\cite{obj1, obj2}, and image segmentation~\cite{seg1, seg2}, primarily driven by the emergence of deep learning. However, the high-performance requirements of these deep learning models have led to their substantial size, resulting in significant computational costs. This poses challenges for practical deployment in real-world industries. To address these limitations, model compression methods, including model pruning~\cite{pruning}, quantization~\cite{quantization}, and knowledge distillation (KD)~\cite{kd_survey}, have been proposed. Among these, KD stands out for its superior performance and ease of implementation, making it widely adopted in various computer vision applications. KD involves training a lightweight student model by distilling meaningful information from a more complex teacher model, enabling the student model to achieve performance similar to that of the teacher model.

\begin{figure}[]
	\centering
	\subfigure[Knowledge distillation with constant temperature scaling \label{fig:const_temperature}]
	{\includegraphics[width=0.8\columnwidth]{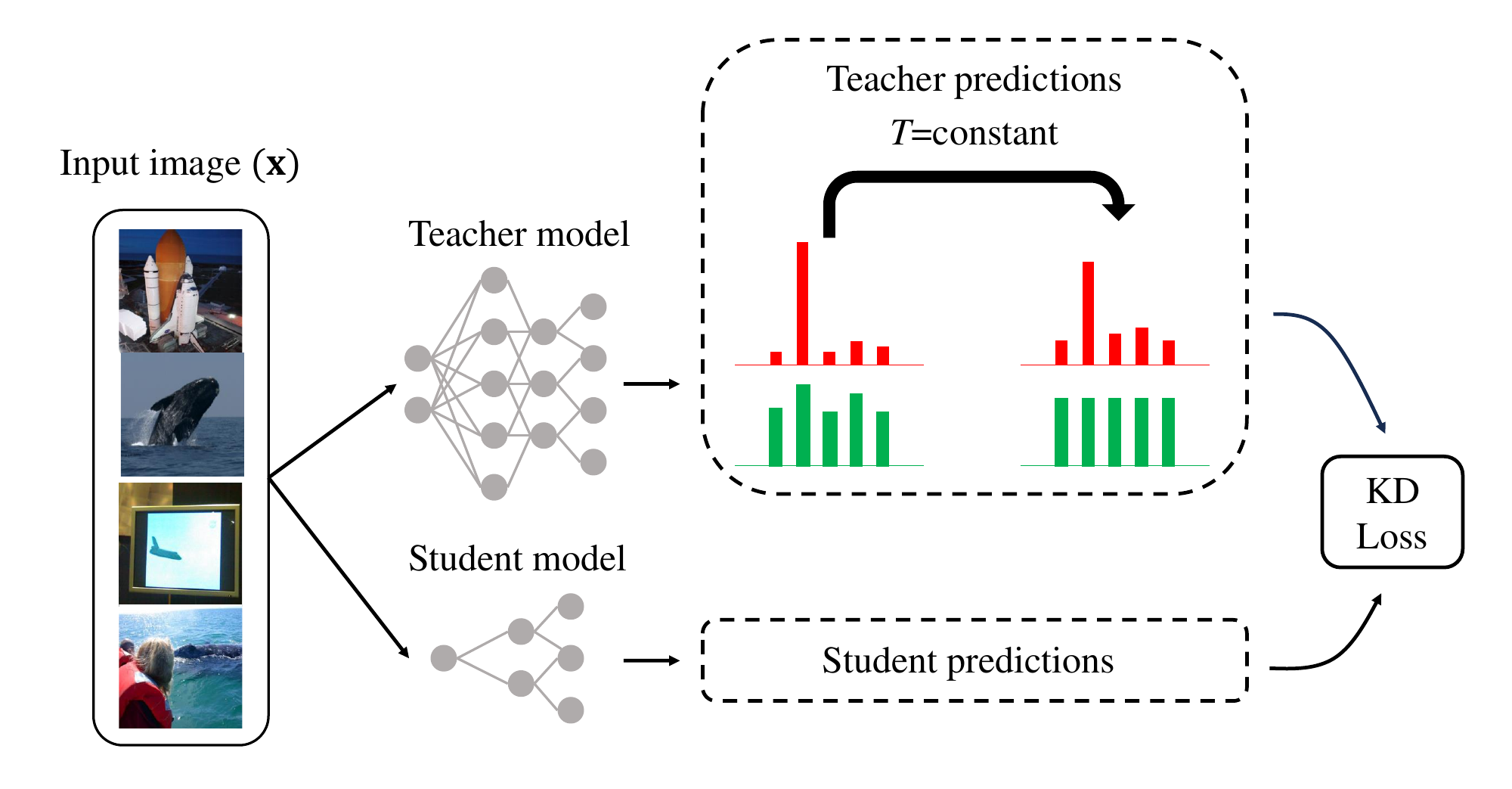}}
	\subfigure[Knowledge distillation with different temperature scaling based on the energy scores \label{fig:ours_temperature}]
	{\includegraphics[width=0.8\columnwidth]{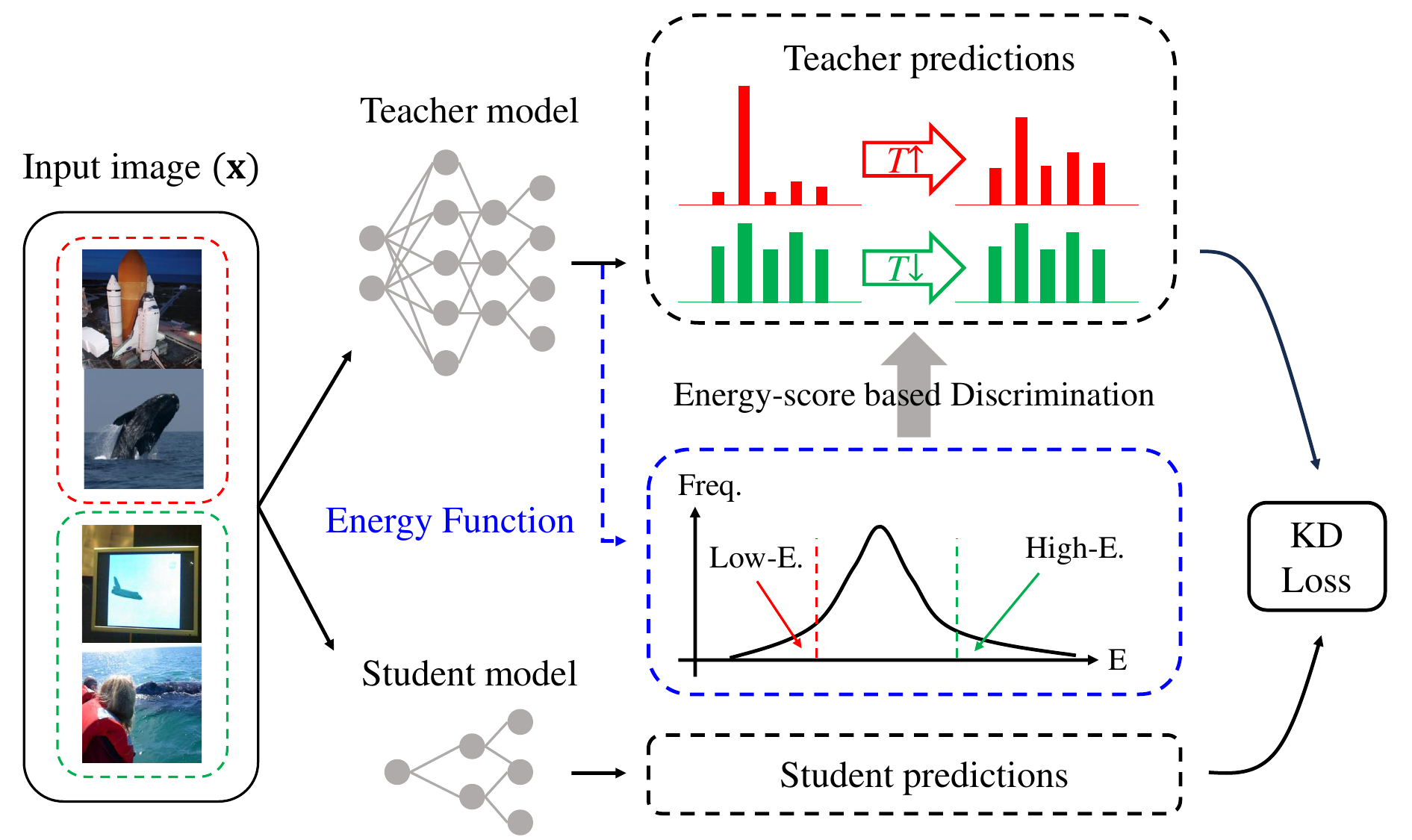}}
	\caption{Schematic diagram of conventional knowledge distillation and our method: (a) constant temperature scaling, (b) different temperature scaling. Our method receives the energy score of each sample from the blue dashed line.}
	\label{fig:overview}
\end{figure}

Since its introduction by Hinton~\cite{hinton}, KD has evolved into two main approaches: logit-based~\cite{dkd} and feature-based~\cite{review} distillation. Logit-based methods use final predictions for training the student, while feature-based methods leverage information from intermediate layers. Although feature-based methods are generally known to outperform logit-based ones, they may be challenging to use in real-world applications due to potential privacy and safety concerns associated with accessing intermediate layers of the teacher model. Hence, this paper focuses on logit-based distillation, offering practical advantages for real-world applications by not requiring access to intermediate layers.

We propose a novel logit-based distillation method designed for easy integration into existing logit-based KDs. This method significantly enhances performance by maximizing the utilization of teacher knowledge by the students. As illustrated in Figure~\ref{fig:overview}, applying an energy function to each image categorizes the entire dataset into low-energy and high-energy samples. We then apply different temperature scaling to the separated samples, employing high temperature for low-energy and low temperature for high-energy samples. This approach results in smoother distributions from low-energy samples and sharper distributions from high-energy samples, effectively adjusting non-target class predictions for optimal knowledge distillation. Consequently, our method significantly improves the performance of the student model.

In addition, we propose High Energy-based Data Augmentation (HE-DA) to further enhance performance. Unlike previous augmentation-based KD methods that apply augmentation to the entire dataset, HE-DA achieves similar or even better performance by utilizing only 20\% to 50\% of the dataset, offering practical advantages in terms of storage and computational cost.

Through extensive experimentation on commonly used classification datasets, such as CIFAR-100~\cite{cifar}, TinyImageNet, and ImageNet~\cite{imagenet}, we have verified that our proposed methods outperform existing state-of-the-art approaches, particularly demonstrating strengths in the case of challenging datasets, such as CIFAR-100-LT and ImageNet.

\section{Related Works}
\label{sec:relatedworks}

\subsection{Knowledge Distillation}
\label{subsec:kd}
Knowledge distillation (KD) is a technique used to enhance the performance of lightweight student networks by leveraging the dark knowledge embedded in large teacher networks. Over the years, KD methods have evolved to narrow the performance gap between student and teacher models by utilizing both final predictions, known as logits-based distillation~\cite{hinton, dml, takd, dkd, jin2023multi}, and intermediate features, known as features-based distillation~\cite{fitnet, pkt, rkd, crd, at, vid, ofd, review, fcfd, cat_kd, lsh-tl, sakd}.

Previous works on logits-based distillations include the following: DML~\cite{dml} proposed a mutual learning strategy for collaboratively teaching and learning between student and teacher models; TAKD~\cite{takd} introduced a multi-step method with an intermediate-size network (i.e., assistant network) to bridge the gap between teachers and students; DKD~\cite{dkd} decomposed the soft-label distillation loss into two components: target class knowledge distillation (TCKD) and non-target class knowledge distillation (NCKD), enabling each part to independently harness its effectiveness; Multi KD~\cite{jin2023multi} proposed multi-level prediction alignment, containing instance, batch, and class levels, and prediction augmentation. While these approaches emphasize effective knowledge transfer, they do not consider dividing entire datasets or provide mechanisms to distinguish and transfer knowledge from specific samples.

FitNet~\cite{fitnet} was groundbreaking as it leveraged not only the final outputs but also intermediate representations. Since the introduction of FitNet, various feature-based KD methods have emerged as follows: AT~\cite{at} prompted the student to mimic the attention map of the teacher network; PKT~\cite{pkt} employed various kernels to estimate the probability distributions, employing different divergence metrics for distillation; RKD~\cite{rkd} focused on transferring the mutual relations of data samples; CRD~\cite{crd} framed the objective as contrastive learning for distillation; VID~\cite{vid} took a different approach by maximizing mutual information; OFD~\cite{ofd} introduced a novel loss function incorporating teacher transform and a new distance function; Review KD~\cite{review} introduced a review mechanism that leverages past features for guiding current ones through residual learning. Additionally, they incorporated attention-based fusion (ABF) and hierarchical context loss (HCL) to further enhance performance.

More recently, several high-performing feature-based distillation methods have emerged~\cite{fcfd, cat_kd, lsh-tl, sakd}. Unlike conventional feature-based methods, which primarily rely on measuring model similarity using isotropic L2 distance, FCFD~\cite{fcfd} focused on optimizing the functional similarity between teacher and student features. By considering the anisotropic nature of neural network operations, FCFD ensured that the student learns more effectively from the teacher. Z. Guo introduced CAT-KD~\cite{cat_kd}, which achieves both high interpretability and competitive performance by transferring class activation maps. This approach is grounded in the understanding that the CNN's ability to distinguish category regions is vital for its performance. J. Li proposed LSH-TL~
\cite{lsh-tl}, which utilizes two algorithms by applying teacher labels and designing feature temperature parameters. This method addresses the challenge of training data not aligning with the ground-truth distribution of classes and features. Lastly, SAKD~\cite{sakd} was presented, employing the sparse attention mechanism by treating student features as queries and teacher features as key values. It utilizes sparse attention values through random deactivation to adjust the feature distances.

While feature-based methods have demonstrated superior performance compared to logits-based ones, attributed to their ability to leverage more information from intermediate layers rather than relying solely on final predictions, they often present challenges in real-world applications. This is due to privacy and safety concerns associated with accessing the intermediate layers of the teacher model. Therefore, in this paper, we shift our focus to logits-based distillation, which offers practical advantages for real-world deployment. To achieve comparable or even superior results to feature-based KDs, we introduce a novel perspective to knowledge distillation: regulating knowledge transfer based on the energy scores of samples.

Our method distinguishes itself from previous KD methods by its compatibility with existing state-of-the-art methods. Unlike previous approaches, which are often limited to standalone use and cannot be easily combined with emerging methods, our approach seamlessly integrates with various methods, providing potential for further performance enhancement. Moreover, our method exhibits significant performance improvements when confronted with challenging samples, rendering it particularly suitable for real-world scenarios. The Energy-based KD (Energy KD) proposed herein represents a significant advancement in the development of more effective and efficient knowledge distillation techniques.

\subsection{Energy-based Learning}
\label{subsec:energy}
Energy-based machine learning models have a long history, beginning with the Boltzmann machine~\cite{ackley1985learning, salakhutdinov2010efficient}, a network of units with associated energy for the entire network. Energy-based learning~\cite{lecun2006tutorial, ranzato2006efficient, ranzato2007unified} offers a unified framework for various probabilistic and non-probabilistic learning approaches. Recent research~\cite{zhao2016energy} demonstrated the use of energy functions to train generative adversarial networks (GANs), where the discriminator utilizes energy values to differentiate between real and generated images. Xie~\cite{xie2018cooperative, xie2018learning, xie2019learning} also established the connection between discriminative neural networks and generative random field models. Subsequent studies have explored the application of energy-based models in video generation and 3D geometric patterns. Liu~\cite{liu2020energy} demonstrated that non-probabilistic energy scores can be directly used in a score function for estimating out-of-distribution (OOD) uncertainty. They show that these optimization goals fit more naturally into an energy-based model than a generative model and enable the exploitation of modern discriminative neural architectures.

Building upon these prior works, our proposed framework extends the use of non-probabilistic energy values to knowledge distillation and data augmentation. Notably, our framework provides different knowledge for low energy and high energy samples, representing a novel contribution.

\section{Methods}
\label{sec:method}

\subsection{Background}
\label{subsec:background}
Our approach revolves around categorizing each sample in the dataset into two groups: low-energy and high-energy groups. These groups are determined by the energy function $E(\cdot)$, which maps the input $\textbf{x}_i$, having dimension $d$, to a single, non-probabilistic scalar value. (i.e., $E(\mathbf{x}_i): \mathbb{R}^d \rightarrow \mathbb{R}$)~\cite{lecun2006tutorial}. The representation of the energy function with the neural network $f$ is as follows:

\begin{equation}
E(\mathbf{x}_i;f)=-T^E \cdot \log \sum_{j=1}^K e^{z^{f}_j(\mathbf{x}_i) / T^E},
\label{eq:energy}
\end{equation}
where $z^f_j(\mathbf{x}_i)$ indicates the logit corresponding to the $j$ class label of input image sample $i$ using the neural network $f$, $T^E$ is the temperature parameter for the energy score, and $K$ denotes the total number of classes. Appendix contains a list of all the symbols mentioned in this paper.

The motivation behind segregating categories according to energy scores is that we can regard input data with low likelihood as high-energy samples~\cite{energyood}. This can be achieved by utilizing the data’s density function $p(\mathbf{x}_i)$ expressed by the energy-based model~\cite{lecun2006tutorial,energy1}.

\begin{equation}   
p(\mathbf{x}_i)=\frac{e^{-E(\mathbf{x}_i ; f) / T^E}}{\int_{\mathbf{x}} e^{-E(\mathbf{x}_i ; f) / T^E}},
\end{equation}
where $\textbf{x}$ denotes the entire dataset and the denominator can be disregarded since it remains constant independently (i.e., $\int_{\mathbf{x}} e^{-E(\mathbf{x}_i ; f) / T^E}= C$). Therefore, it can be expressed by 

\begin{equation}
   \log p(\mathbf{x}_i)=-\frac{E(\mathbf{x}_i ; f)}{T^E}-\log C.
\label{eq:prob}
\end{equation}

This equation shows that the energy function is proportional to the log likelihood function. In other words, samples with lower energy have a higher probability of occurrence, indicating a \textit{certain image}, while samples with higher energy have a lower probability of occurrence, referring a \textit{uncertain image}. This distinguishable nature of the energy function can be effectively utilized to categorizes samples, thereby facilitating optimal knowledge distillation. To visually explore the relationship between energy scores and image, refer to Figure~\ref{fig:visualization}, which shows images associated with low and high energy, respectively. Consequently, the energy score, being a valuable tool for dataset division, can be employed in both knowledge distillation (KD) and data augmentation (DA) separately. Further elaboration on each method will be provided in the subsequent sections.

\subsubsection{Low and High Energy Samples}
\label{subsec:energysamples}

\begin{figure}[h]
	\centering
	\subfigure[Energy scores over all samples\label{fig:energyscore}]
	{\includegraphics[width=0.45\columnwidth]{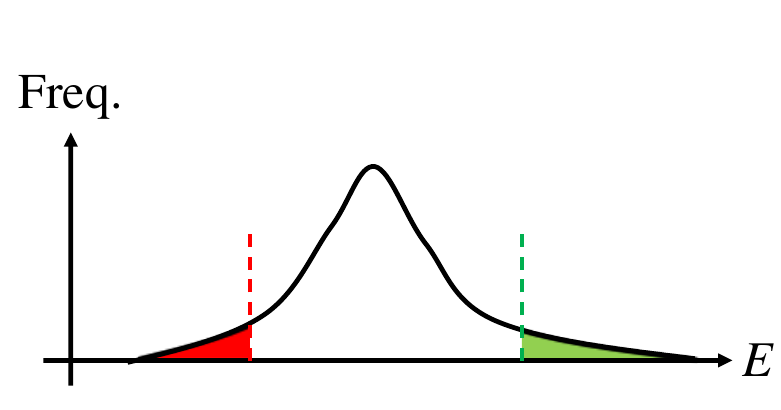}}
	\subfigure[Electric ray\label{fig:electricray}]
	{\includegraphics[width=0.45\columnwidth]{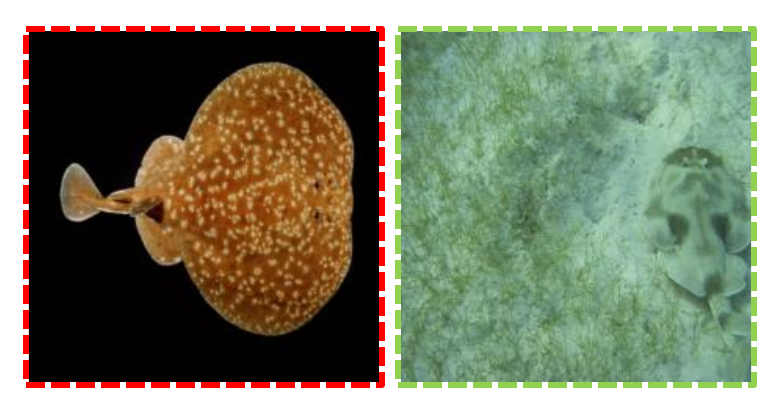}}
 	\subfigure[Loggerhead\label{fig:energyscore}]
	{\includegraphics[width=0.45\columnwidth]{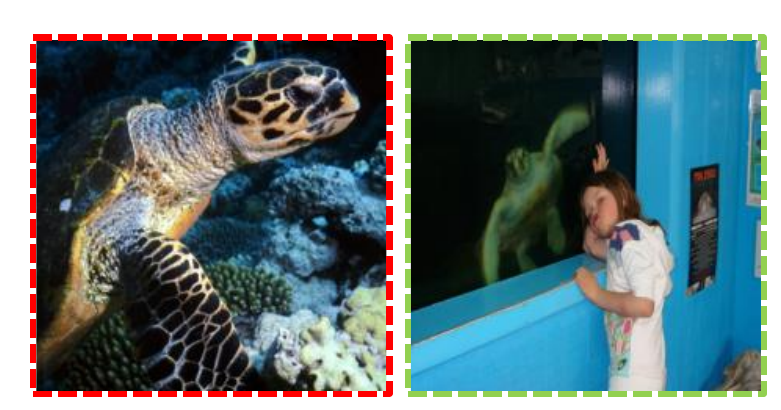}}
	\subfigure[Respirator\label{fig:electricray}]
	{\includegraphics[width=0.45\columnwidth]{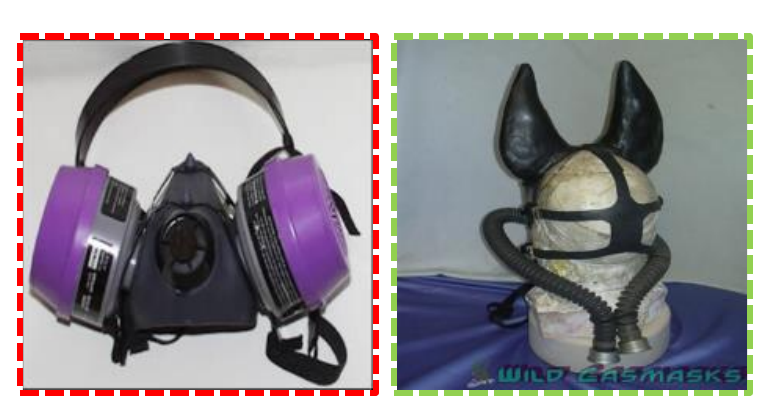}}
	\caption{ImageNet samples categorized according to their energy scores obtained from ResNet32x4. The red boxes belong to the certain images and have low energy scores, accurately representing their assigned labels. The green boxes are relative to the uncertain images and have high energy scores, not clearly reflecting their assigned labels.}
	\label{fig:visualization}
\end{figure}

To validate the insights gained from the energy scores, it is valuable to visualize the images belonging to both the low-energy and high-energy.

\begin{figure}[]
	\centering
	\subfigure[Electric ray\label{fig:energyscore}]
	{\includegraphics[width=0.75\columnwidth]{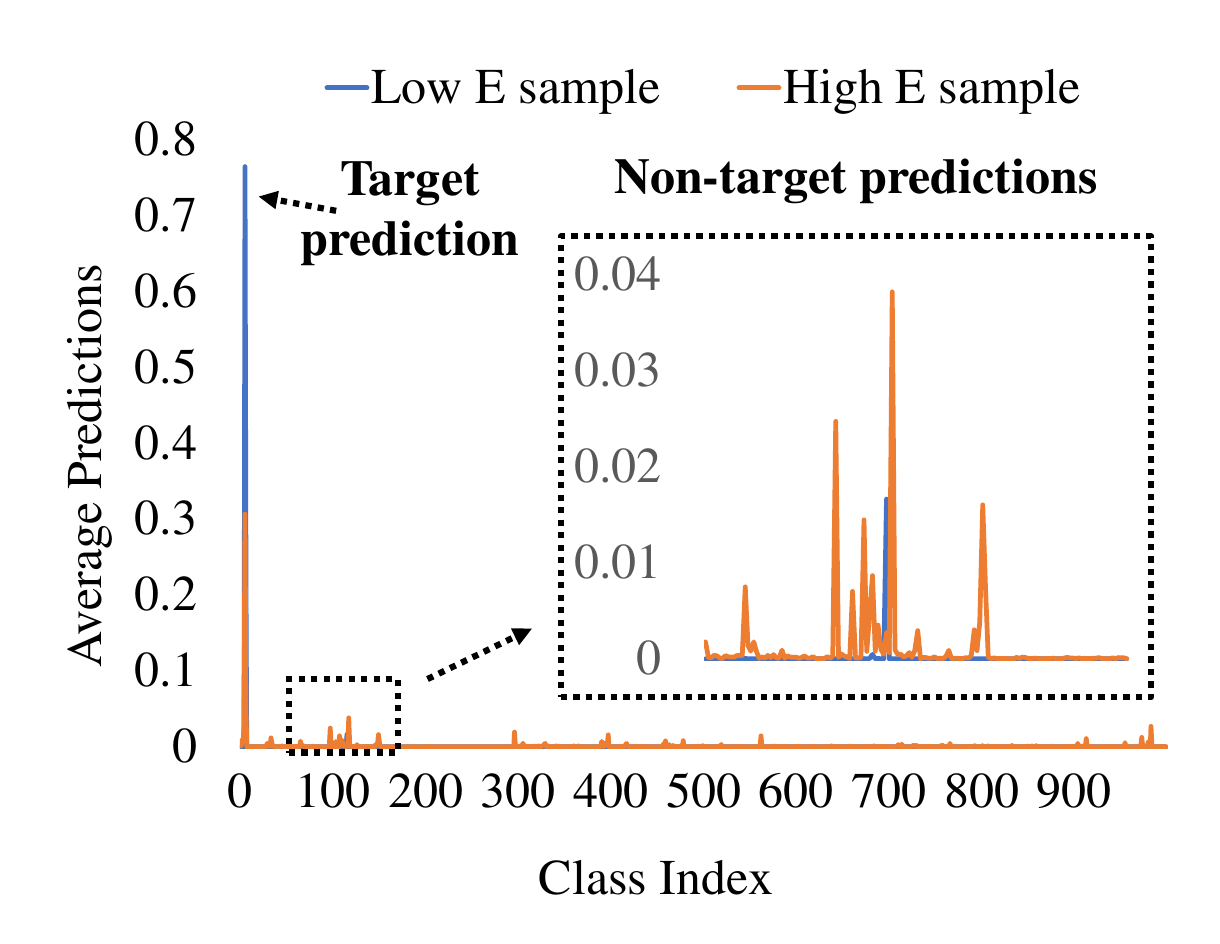}}
	\subfigure[Loggerhead\label{fig:electricray}]
	{\includegraphics[width=0.75\columnwidth]{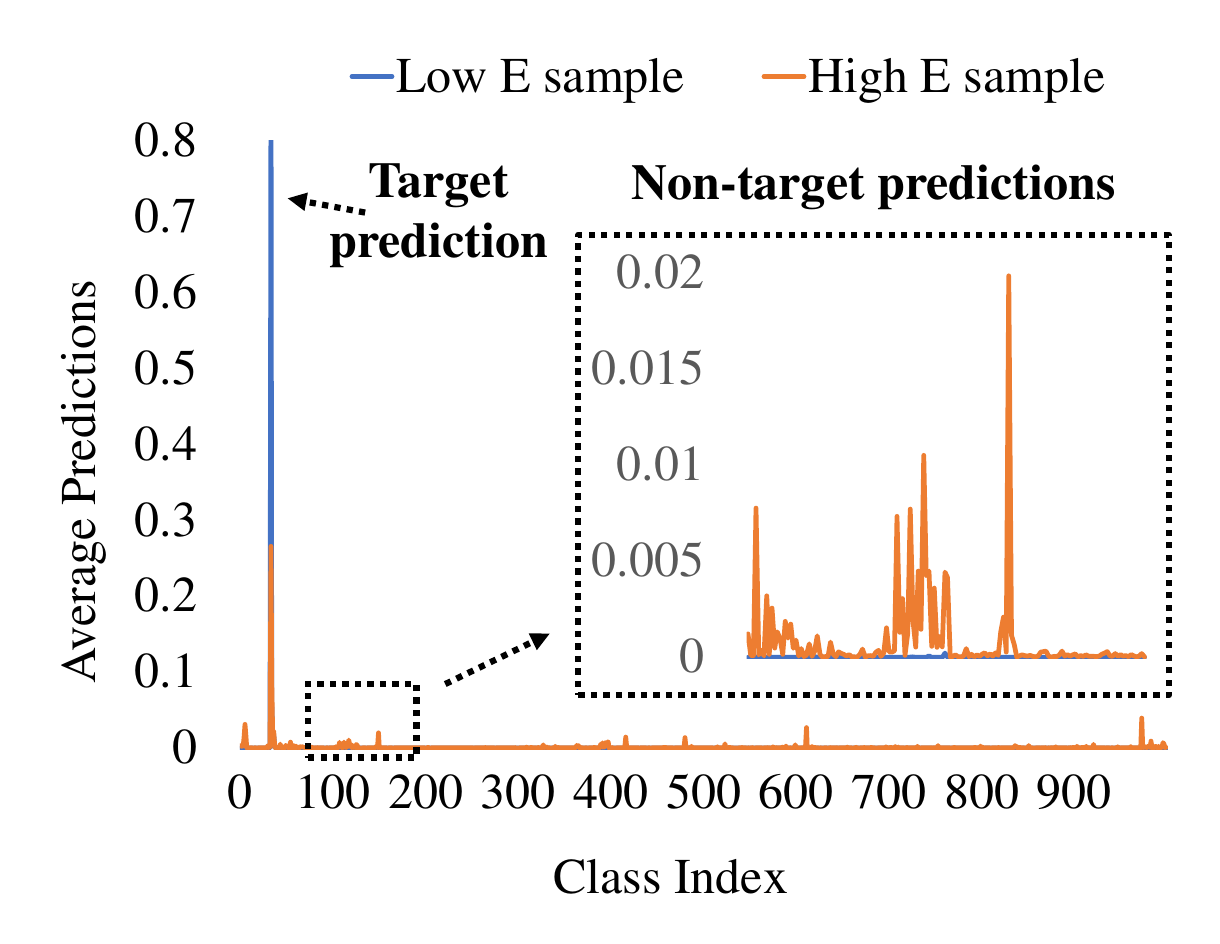}}
 	\subfigure[Respirator\label{fig:energyscore}]
	{\includegraphics[width=0.75\columnwidth]{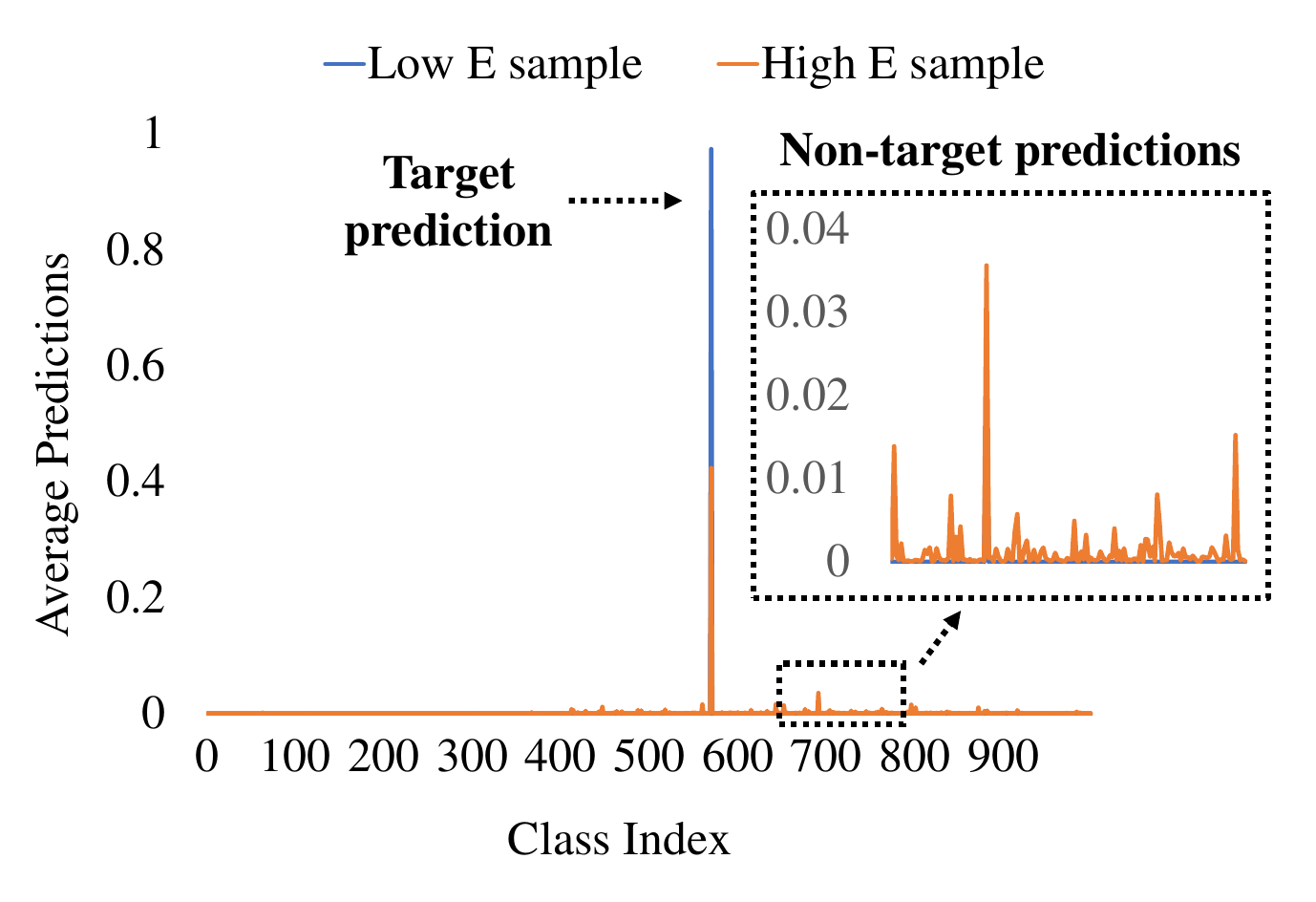}}
	\caption{Average predictions for particular classes with low energy (blue line) and high energy (red line) samples. Low energy samples exhibit high confidence scores and lack substantial dark knowledge, whereas high energy samples display low confidence scores and have inordinate knowledge.}
	\label{fig:predictions}
\end{figure}

Figure~\ref{fig:visualization} illustrates the categorization of ImageNet based on the energy score of each image, dividing them into categories of low-energy and high-energy samples. The red boxes depicts images with low-energy scores, effectively representing their respective classes. We have denoted this category as \textit{certain images}. On the other hand, the green boxes displays images with high-energy scores, indicating either a confused label or a mixture of different objects. These images have been designated as \textit{uncertain images}. 

Figure~\ref{fig:predictions} demonstrates the average predictions for \textit{certain} and \textit{uncertain} images. Certain images exhibit high confidence scores and possess insufficient knowledge about non-target predictions, while uncertain images showcase low confidence scores and a relatively uniform distribution. It's worth noting that the predictions presented in Figure~\ref{fig:predictions} support the classification of each image as either certain or uncertain. These findings align with prior research~\cite{energyood} that higher energy levels are associated with out-of-distribution (OOD) data. An important difference is that we categorize low-energy and high-energy data within the same dataset based on our criteria. Additional images are available in Appendix.

As a result, it is reasonable to utilize higher temperature scaling for low-energy samples to create smoother predictions and lower temperature scaling for high-energy samples to achieve sharper predictions during the distillation process. This ensures that the teacher model optimally transfers its knowledge to the student model.

\subsection{EnergyKD: Energy-based Knowledge Distillation}
Utilizing the mentioned energy score, we propose an Energy-based Knowledge Distillation (Energy KD), where the differences between low and high energy allow effective transfer of knowledge. Specifically, we obtain the energy score for each image sample through the logits of pre-trained teacher models using Eq.~\ref{eq:energy}. After classifying the images into low-energy and high-energy groups based on their energy scores, we apply distinct softmax temperature scaling to each group, thereby enhancing the student model's ability to learn a more diverse range of information.

First of all, we consider teacher network $f_\mathcal{T}$ and student network $f_\mathcal{S}$, which maps input image $\textbf{x}_i$ with dimension $d$ to number of classes $K$ (i.e., $f_\mathcal{T, S}: \mathbb{R}^d \rightarrow \mathbb{R}^K$) as follows:
\begin{equation}
f_\mathcal{T} = f_\mathcal{T}(\mathbf{x}_i; \theta_\mathcal{T})
\end{equation}
\begin{equation}
    f_\mathcal{S} = f_\mathcal{S}(\mathbf{x}_i; \theta_\mathcal{S})
\end{equation}

Here, $\theta_\mathcal{T}$ and $\theta_\mathcal{S}$ are the parameters of each model. The energy score $\mathcal{E}$ for each sample $\mathbf{x}_i$ can be calculated using the teacher network $f_\mathcal{T}$ as follows:

\begin{equation}
        \mathcal{E}^{(i)}_\mathcal{T} = E(\mathbf{x}_i ; f_\mathcal{T}).
\label{eq:eq6}
\end{equation}

We can obtain the energy score for all images in the training dataset and arrange them in ascending order. Subsequently, images with lower energy values are categorized into the certain group, while those with higher energy values are assigned to the uncertain group.

In contrast to the conventional KD loss $\mathcal{L}_\mathrm{KD}$, which employs the same temperature scaling for all images, as indicated by

\begin{equation}
\mathcal{L}_\mathrm{KD}(\mathbf{x}_i;f_\mathcal{S},f_\mathcal{T},T) = \mathcal{D}_\mathrm{KL}\Big(\sigma\Big(\frac{z^{f_\mathcal{T}}}{T}\Big)\Big|\Big|\sigma\Big(\frac{z^{f_\mathcal{S}}}{T}\Big)\Big),  
\label{eq:general-KLD}
\end{equation}where $\mathcal{D}_\mathrm{KL}$ denotes Kullback–Leibler divergence, $\sigma(\cdot)$ is the softmax function, $T$ is the temperature scaling factor, and $z^{f_\mathcal{T}}, z^{f_\mathcal{S}}$ indicate the logit using the teacher network $f_\mathcal{T}$, student network $f_\mathcal{S}$, respectively. 

\begin{figure}[t]
\centering
\includegraphics[width=1.0\columnwidth]{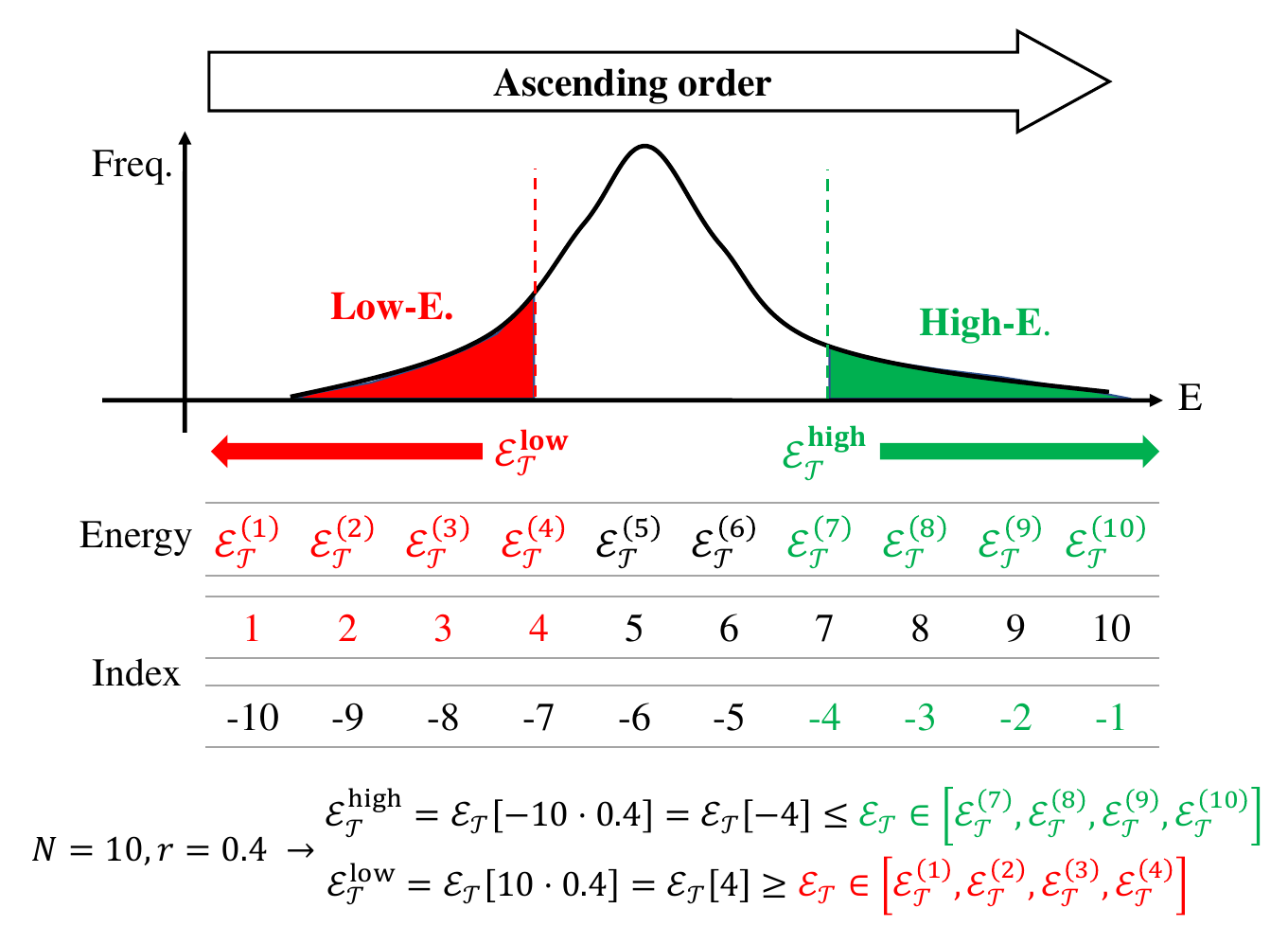}
\caption{Energy distribution across the entire datasets. This illustrated example assumes that there are 10 image samples and sets the percentage of the total samples to 40\%.}
\label{fig:array}
\end{figure}

Our method adjusts the confidence of predictions based on the energy score, enabling the student to acquire a broader range of knowledge. This adjustment can be utilized by simply changing the temperature scaling factor ($T \rightarrow T_\mathrm{ours}$) as follows:

\begin{equation}
\mathcal{L}_\mathrm{ours}(\mathbf{x}_i;f_\mathcal{S},f_\mathcal{T},T_\mathrm{ours}) = \mathcal{D}_\text{KL}\Big(\sigma\Big(\frac{z^{f_\mathcal{T}}}{T_\mathrm{ours}}\Big)\Big|\Big|\sigma\Big(\frac{z^{f_\mathcal{S}}}{T_{\mathrm{ours}}}\Big)\Big)
\label{eq:ours}
\end{equation}

\begin{equation}
T_\mathrm{ours} = \begin{cases} T + T_{(-)}, & \mathcal{E}^{(i)}_\mathcal{T} \geq \mathcal{E}_\mathcal{T}^\mathrm{high} = \mathcal{E}_\mathcal{T}[-N\cdot r]
\\ T + T_{(+)}, & \mathcal{E}^{(i)}_\mathcal{T} \leq \mathcal{E}_{\mathcal{T}}^\mathrm{low} = \mathcal{E}_\mathcal{T}[N\cdot r] 
\\ T, & \mathrm{else}
\end{cases},
\label{eq:eq9}
\end{equation}where $\mathcal{E}_{\mathcal{T}}^\mathrm{high}$ and $\mathcal{E}_{\mathcal{T}}^\mathrm{low}$ are constant values that define the range of high-energy and low-energy classifications. $T_{(-)}$ is a negative integer used to decrease the temperature, facilitating the transfer of more target predictions for uncertain samples. Conversely, $T_{(+)}$ is a positive integer used to increase the temperature, enabling certain samples to incorporate more non-target knowledge. $N$ represents the total number of training samples, and to establish the range based on energy, we employed a percentage of the total samples denoted by $r=\{0.2, 0.3, 0.4, 0.5\}$. The parentheses $[\cdot]$ indicate the index of the array, as shown in Figure~\ref{fig:array} for a more intuitive understanding.

Figure~\ref{fig:array} illustrate how our method divides the entire dataset into low-energy and high-energy samples. To facilitate understanding, we employ a dataset comprising 10 samples and set the percentage parameter $r$ to $0.4$. Eq.~\ref{eq:eq6} calculates the energy score for each data sample, and then we arrange in ascending order (i.e., $\mathcal{E}^{(1)}_\mathcal{T}<\mathcal{E}^{(2)}_\mathcal{T}<...<\mathcal{E}^{(10)}_\mathcal{T}$). To aid comprehension, we introduce two different indexes: negative and positive indexes. As shown in Eq.~\ref{eq:eq9}, $\mathcal{E}^\mathrm{high}_\mathcal{T}$ is set to $\mathcal{E}_\mathcal{T}\left[-4\right]$ and $\mathcal{E}^\mathrm{low}_\mathcal{T}$ is set to $\mathcal{E}_\mathcal{T}\left[4\right]$. Consequently, samples with energy score $\mathcal{E}_\mathcal{T}\left[-4\right], \mathcal{E}_\mathcal{T}\left[-3\right], \mathcal{E}_\mathcal{T}\left[-2\right]$, and $\mathcal{E}_\mathcal{T}\left[-1\right]$ (equivalent to $\mathcal{E}_\mathcal{T}\left[7\right], \mathcal{E}_\mathcal{T}\left[8\right], \mathcal{E}_\mathcal{T}\left[9\right]$, and $\mathcal{E}_\mathcal{T}\left[10\right]$), equal to or greater than $\mathcal{E}_\mathcal{T}\left[-4\right]$, belong to high-energy samples, Conversely, samples with energy score $\mathcal{E}_\mathcal{T}\left[1\right], \mathcal{E}_\mathcal{T}\left[2\right], \mathcal{E}_\mathcal{T}\left[3\right]$, and $\mathcal{E}_\mathcal{T}\left[4\right]$, equal to or less than $\mathcal{E}_\mathcal{T}\left[4\right]$, belong to low-energy sample.

Hinton's paper~\cite{hinton} introducing the concept of KD softened the probabilities by setting the temperature used for softmax higher than normal ($T=1$). These relative probabilities of incorrect answers provide valuable insights into the generalization tendencies of the complex model~\cite{hinton}. Since Hinton’s work, this information in soft targets has been termed \textit{dark knowledge}~\cite{dark1, dark2, kd_survey}. Following this line of research, we also adopted the term \textit{dark knowledge} to grasp the useful probability information. Previous research indicated that, in order to enhance the performance of the KD, \textit{dark knowledge} must be appropriately distributed~\cite{ATS}. Our approach can increase important \textit{dark knowledge} about non-target classes in the low energy sample while increasing predictions of the target class in the high-energy sample.

\begin{table*}[h]
\centering
\resizebox{0.6\textwidth}{!}{%
\begin{tabular}{c|cccccc}
\toprule
Teacher    & WRN-40-2                              & WRN-40-2                             & ResNet56                          & ResNet110                                & ResNet32x4                                & VGG13                           \\
Acc.       & 75.61                                 & 75.61                                 & 72.34                                 & 74.31                                 & 79.42                                 & 74.64                                 \\
Student    & WRN-16-2                              & WRN-40-1                             & ResNet20                            & ResNet32                                 & ResNet8x4                           & VGG8                          \\
Acc.       & 73.26                                 & 71.98                                 & 69.06                                 & 71.14                                 & 72.50                                 & 70.36                                 \\ \midrule
FitNet     & 73.58                                 & 72.24                                 & 69.21                                  & 71.06                                 & 73.50                                 & 71.02                                 \\
PKT        & 74.54                                 & 73.54                                 & 70.34                           & 72.61                                 & 73.64                                 & 72.88                           \\
RKD        & 73.35                                 & 72.22                                 & 69.61                                  & 71.82                                 & 71.90                                 & 71.48                                 \\
CRD        & 75.48                                 & 74.14                                 & 71.16                                  & 73.48                                 & 75.51                                 & 73.94                                 \\
AT         & 74.08                                 & 72.77                                 & 70.55                                 & 72.31                                 & 73.44                                 & 71.43                                 \\
VID        & 74.11                                 & 73.30                                 & 70.38                                 & 72.61                                 & 73.09                                 & 71.23                                 \\
OFD        & 75.24                                 & 74.33                                 & 70.98                                 & 73.23                                 & 74.95                                 & 73.95                                 \\
ReviewKD        & 76.12                                 & 75.09                                 & 71.89                                 & 73.89                                 & 75.63                                 & 74.84                                 \\
FCFD        & 76.34                                 & 75.43                                 & 71.68                                 & -                                & 76.80                                 & 74.86                                 \\
CAT-KD        & 75.60                                 & 74.82                                 & 71.62                                 & 73.62                                 & 76.91                                 & 74.65                                 \\
LSH-TL        & 76.42                                 & 74.50                                 & 71.50                                 & 74.10                                 & 76.73                                 & 74.11                                 \\
SAKD        & 75.86                                 & 75.00                                 & 71.93                                 & 73.92                                 & 76.16                                 & 74.66                                 \\\midrule
DML         & 73.58                         & 72.68                          & 69.52                                 & 72.03                           & 72.12                           & 71.79                                 \\
TAKD         & 75.12                         & 73.78                          & 70.83                                 & 73.37                           & 73.81                           & 73.23                                 \\ \midrule
KD         & 74.92                           & 73.54                           & 70.66                                 & 73.08                           & 73.33                           & 72.98                                 \\ 
\textbf{Energy KD}  & \textbf{75.45}                        & \textbf{74.28}                        & \textbf{71.30}                        & \textbf{73.68}                        & \textbf{74.60}                        & \textbf{73.73}                        \\
\midrule
DKD         & 76.24                         & 74.81                          & 71.97                                 & 74.11                           & 76.32                           & 74.68                                 \\
\textbf{Energy DKD}  & \textbf{76.66}                        & \textbf{74.97}                        & \textbf{72.10}                        & \textbf{74.11}                        & \textbf{76.78}                        & \textbf{74.90}                        \\
\midrule
Multi KD         & 76.63                         & 75.35                          & 72.19                                 & 74.11                           & 77.08                           & 75.18                                 \\ 
\textbf{Energy Multi}  & \textbf{77.19}                        & \textbf{75.70}                        & \textbf{72.76}                        & \textbf{74.60}                        & \textbf{77.31}                        & \textbf{75.56}                        \\
\bottomrule
\end{tabular}%
} 
\caption{Top-1 accuracy (\%) on the CIFAR-100 test sets when using teacher and student models with the same architectures. Our results, highlighted in \textbf{bold}, demonstrate exceptional performance compared to the results obtained without employing our method.}
\label{table:cifar_same}
\end{table*}

\subsection{HE-DA: High Energy-based Data Augmentation}
\label{subsec:he-da}
We propose an additional technique, High Energy-based Data Augmentation (HE-DA), where data augmentation is selectively applied only to image samples belonging to high-energy groups, which have already been classified for Energy KD. In conventional knowledge distillation (KD), data augmentation (DA) is frequently employed across the entire dataset to enhance the generalization and performance of the student model. However, the straightforward application of DA may result in a significant increase in computational costs due to the doubling of the dataset.

To efficiently apply DA to KD, we present an augmentation method that focuses on specific samples (i.e., uncertain samples) instead of augmenting the entire dataset. This approach is rooted in the concept that certain samples already contain sufficient information, whereas uncertain samples require additional information to elucidate ambiguous content. Eq.~\ref{eq:prob} and Figure~\ref{fig:visualization} demonstrates that high-energy samples correspond to uncertain samples. Consequently, our focus is directed towards augmenting the samples within the high-energy group, aiming to provide students with more information to enhance their performance.

In our Energy KD approach, we sorted the energy scores obtained from the teacher model in ascending order. The samples with lower values are categorized as part of the low-energy dataset $\textbf{x}_\mathrm{low}$ within the entire dataset $\textbf{x}$, while the samples with higher values are classified as belonging to the high-energy dataset $\textbf{x}_\mathrm{high}$ as follows:

\begin{equation}
    \mathbf{x} = \{\mathbf{x}_\mathrm{low}, \mathbf{x}_\mathrm{else},  \mathbf{x}_\mathrm{high}\} 
\end{equation}

\begin{equation}
\mathbf{x}_i = \begin{cases}\mathbf{x}_{\mathrm{high}}, & \mathcal{E}^{(i)}_\mathcal{T} \geq \mathcal{E}^\mathrm{high}_\mathcal{T} = \mathcal{E}_\mathcal{T}[-N\cdot r] \\ \mathbf{x}_{\mathrm{low}}, & \mathcal{E}^{(i)}_\mathcal{T} \leq \mathcal{E}^\mathrm{low}_\mathcal{T} = \mathcal{E}_\mathcal{T}[N\cdot r]
\\
\mathbf{x}_{\mathrm{else}}, & \mathrm{else}\end{cases},
\end{equation}
where $\textbf{x}_\mathrm{else}$ represents datasets that do not belong to either low-energy or high-energy data and the parentheses $[\cdot]$ denote the index of the array, as mentioned earlier. For a clearer understanding, please refer back to Figure~\ref{fig:array}. We exclusively apply augmentation to samples that were classified as part of the high energy $\textbf{x}^\mathrm{aug}_\mathrm{high}$ as follows:

\begin{equation}
    \mathbf{x}_\mathrm{high}^\mathrm{aug} = G_\mathrm{aug}(\mathbf{x}_\mathrm{high}),
\end{equation}
where $G_\mathrm{aug}$ refers the data augmentation function, here, we applied CutMiX~\cite{cutmix} and MixUp~\cite{mixup}.

Despite utilizing only an additional 20\% to 50\% of the training data, our method outperforms existing approaches, which use the entire dataset, yielding superior results while simultaneously reducing computational costs. It is noteworthy that our approach demonstrates higher performance than applying data augmentation only to low-energy samples or applying data augmentation to both low and high-energy samples. Please refer to Section~\ref{subsec:he-da_contribution} for more details. The results with MixUp are included in Appendix.

\section{Experiments}
\label{sec:experiment}

\begin{table*}[h]
\centering
\resizebox{0.65\textwidth}{!}{%
\begin{tabular}{c|cccccc}
\toprule
Teacher    & WRN-40-2                              & ResNet50                            & ResNet32x4                         & ResNet32x4                                & VGG13                                                           \\
Acc.       & 75.61                                 & 79.34                                 & 79.42                                 & 79.42                                 & 74.64                                                                  \\
Student    & ShuffleNetV1                              & MobileNetV2                             & ShuffleNetV1                            & ShuffleNetV2                                 & MobileNetV2                                                    \\
Acc.       & 70.50                                 & 64.60                                 & 70.50                                 & 71.82                                 & 64.60                                                                  \\ \midrule
FitNet     & 73.73                                 & 63.16                                 & 73.59                                  & 73.54                                 & 64.14                                                                  \\
PKT        & 73.89                                 & 66.52                                 & 74.10                           & 74.69                                 & 67.13                                                            \\
RKD        & 72.21                                 & 64.43                                 & 72.28                                  & 73.21                                 & 64.52                                                                 \\
CRD        & 76.05                                 & 69.11                                 & 75.11                                  & 75.65                                 & 69.73                                                                 \\
AT         & 73.32                                 & 58.58                                 & 71.73                                 & 72.73                                 & 59.40                                                                  \\
VID        & 73.61                                 & 67.57                                 & 73.38                                 & 73.40                                 & 65.56                                                                  \\
OFD        & 75.85                                 & 69.04                                 & 75.98                                 & 69.82                                 & 68.48                                                                 \\
ReviewKD        & 77.14                                 & 69.89                                 & 77.45                                 & 77.78                                 & 70.37                                 &                                \\
FCFD        & 77.81                                 & 71.07                                 & 78.12                                 & 78.20                                 & 70.67                                 &                                \\
CAT-KD        & 77.35                                 & 71.36                                  & 78.26                                  & 78.41                                  & 69.13                                  &                                \\
LSH-TL        & 76.62                                 & 68.02                                 & 75.80                                 & 76.62                                 & 68.17                                 &                                \\
SAKD        & 76.77                                 & -                                 & 76.32                                 & 77.21                                 & -                                &                                \\\midrule
DML         & 72.76                         & 65.71                          & 72.89                                 & 73.45                           & 65.63                                                           \\
TAKD         & 75.34                         & 68.02                          & 74.53                                 & 74.82                           & 65.63                                                           \\ \midrule
KD         & 74.83                           & 67.35                           & 74.07                                 & 74.45                           & 67.37                                                            \\
\textbf{Energy KD}  & \textbf{75.90}                        & \textbf{68.97}                        & \textbf{75.20}                        & \textbf{75.87}                        & \textbf{68.65}                                               \\ 
\midrule
DKD         & 76.70                         & 70.35                          & 76.45                                 & 77.07                           & 69.71                                                            \\ 
\textbf{Energy DKD}  & \textbf{77.06}                        & \textbf{70.77}                        & \textbf{76.89}                        & \textbf{77.55}                        & \textbf{70.19}                                              \\
\midrule
Multi KD         & 77.44                         & 71.04                          & 77.18                                 & 78.44                           & 70.57                                                           \\ 
\textbf{Energy Multi}  & \textbf{77.76}                        & \textbf{71.32}                        & \textbf{77.82}                        & \textbf{78.64}                        & \textbf{70.89}                                                \\
\bottomrule
\end{tabular}%
} 
\caption{Top-1 accuracy (\%) on the CIFAR-100 test sets when using teacher and student models with the different architectures. Our results, highlighted in \textbf{bold}, demonstrate exceptional performance compared to the results obtained without employing our method.}
\label{table:cifar_diff}
\end{table*}

\begin{table*}[]
\centering
\resizebox{0.65\textwidth}{!}{%
\begin{tabular}{ccc|cccc|ccc}
\toprule
\multicolumn{3}{c|}{Distillation} & \multicolumn{4}{c|}{Features}    & \multicolumn{3}{c}{Logits} \\ \midrule
   R50-MV1 & Teacher      & Student     & AT    & OFD   & CRD   & ReviewKD & KD           & DKD  & \textbf{Energy DKD}         \\ \midrule
   Top-1      & 76.16        & 68.87       & 69.56 & 71.25 & 71.37 & \underline{72.56}    & 68.58       & 72.05   & 
 \textbf{72.98}       \\
Top-5      & 92.86        & 88.76       & 89.33 & 90.34 & 90.41 & 91.00    & 88.98    &   \underline{91.05} &   \textbf{91.31}  

\\ \midrule
 R34-R18& Teacher      & Student     & AT    & OFD   & CRD   & ReviewKD & KD     & DKD  & \textbf{Energy DKD}         \\ \midrule
Top-1      & 73.31        & 69.75       & 70.69 & 70.81 & 71.17 & 71.61    & 70.66 &  \underline{71.70}   & \textbf{72.21}      \\
Top-5      & 91.42        & 89.07       & 90.01 & 89.98 & 90.13 & \underline{90.51}    & 89.88 & 90.41  & \textbf{90.81} \\
\bottomrule
\end{tabular}
}

\caption{{Top-1 and Top-5 accuracy~(\%) on the ImageNet validation.} In the row above, ResNet50 is the teacher and MobileNetV1 is the student. In the row below, ResNet34 is the teacher and ResNet18 is the student. The best results are highlighted in \textbf{bold} and the second best \underline{underlined}.}

\label{Table:ImageNet}
\end{table*}

The performance of our method is evaluated by comparing it to previous knowledge distillations such as KD~\cite{hinton}, AT~\cite{at}, OFD~\cite{ofd}, CRD~\cite{crd}, FitNet~\cite{fitnet}, PKT~\cite{pkt}, RKD~\cite{rkd}, VID~\cite{vid}, DML~\cite{dml}, TAKD~\cite{takd}, DKD~\cite{dkd}, ReviewKD~\cite{review}, Multi KD~\cite{jin2023multi}, FCFD~\cite{fcfd}, CAT-KD~\cite{cat_kd}, LSH-TL~\cite{lsh-tl}, and SAKD~\cite{sakd} considering various architectural configurations including ResNet~\cite{resnet}, WideResNet~\cite{wideresnet}, VGG~\cite{vgg}, MobileNet~\cite{mobilenet}, and ShuffleNet~\cite{shufv1, shufv2}. Details of the implementation can be found in Appendix. \textit{All experiments were conducted three times, and the reported results represent the average values.}

\subsection{Datasets}
\label{subsec:datasets}
\textbf{CIFAR-100}~\cite{cifar} is a widely used dataset for image classification, consisting of 100 classes. The samples have a resolution of $32\times32$ pixels, and the dataset includes 50,000 training images and 10,000 testing images.

\textbf{ImageNet}~\cite{imagenet} is a comprehensive dataset extensively employed for image classification. It comprises 1,000 classes, and the samples are of size $224\times224$ pixels. The training set is notably large, containing 1.28 million images, while the test set consists of 5,000 images.

\textbf{TinyImageNet} is a scaled-down version of ImageNet, featuring 200 classes with images sized $64\times64$ pixels. The dataset includes 500 training images, 50 validation images, and 50 testing images for each class.
 
\subsection{Effect of EnergyKD}
\label{subsec:energykdeffect}

\begin{table*}[]
\centering
\resizebox{0.7\textwidth}{!}{%
\begin{tabular}{ccc||ccccc}
\toprule
Teacher       & ResNet32x4     & ResNet32x4     & Teacher       & WRN-40-2       & ResNet32x4     & VGG 13         & ResNet32x4     \\
Student       & ResNet8x4      & ShuffleNetV2   & Student       & WRN-16-2       & ResNet8x4      & MobileNetV2    & ShuffleNetV2   \\ \midrule
Low-          & 73.27          & 75.38          & KD            & 75.06          & 73.33          & 67.37          & 74.45          \\
High-         & 73.88          & 75.34          & Gradation     & 75.49          & 74.32          & 68.57          & 75.74          \\ \midrule
\textbf{Ours} & \textbf{74.60} & \textbf{75.87} & \textbf{Ours} & \textbf{75.45} & \textbf{74.60} & \textbf{68.65} & \textbf{75.87} \\
\bottomrule
\end{tabular}%
}
\caption{Left: Performance evaluated based on the sample type. 'Low-' indicates the application of temperature scaling only to low energy samples (i.e., high T), while 'High-' signifies the utilization of temperature scaling solely for high energy samples (i.e., low T)., Right: Comparing the effectiveness of temperature gradation with the temperature utilized in the performance analysis of our approach.}
\label{tab:energykd_ablations}
\end{table*}

Table~\ref{table:cifar_same} displays the results obtained using the same architecture for both teacher and student models on the CIFAR-100 dataset, while Table~\ref{table:cifar_diff} showcases the results obtained with different architectures. Previous methods can be categorized into two types: feature-based methods and logit-based methods, and the results from previous papers on each method are recorded.

The tables consistently demonstrate that the application of our method to previous logit-based KD results in higher performance compared to not applying it, regardless of structural differences between the student and teacher models. In the case of vanilla KD, our method (Energy KD) yields a performance gain of up to 1.6.

Additionally, when integrating our method into recently developed logit-based methods like DKD and Multi KD (Energy DKD and Energy Multi), we observe performance gains of up to 0.5 and 0.6, respectively. Notably, our method, despite being logit-based, outperforms the state-of-the-art feature-based KD for all same architectures. Additionally, for different architectures, we observe that our method yields results almost comparable to those of the state-of-the-art feature-based methods.

These results suggest that our method holds the potential for seamless integration into future logit-based methods, providing a pathway to further enhance performance. This underscores the superiority of our method for real-world applications, particularly in scenarios where utilizing an intermediate layer is challenging.

Table~\ref{Table:ImageNet} presents the performance of our methods on ImageNet. Notably, even on ImageNet, considered a more challenging dataset than CIFAR-100, our method demonstrates significant improvements over other distillation methods. This improvement is attributed to the optimization of knowledge distillation for this challenging dataset, achieved by applying different temperatures to high- and low-energy samples based on the energy score of the images. On a Top-1 basis, our method achieved a performance improvement of up to 0.6\% over ReviewKD and up to 0.93\% over DKD. For detailed hyperparameter settings, please refer to Appendix.

\subsection{Temperature Ablations}
\label{subsec:temperaturealbation}

Earlier, we applied different temperatures to low-energy and high-energy samples. To assess the feasibility of employing distinct temperature scaling for both energy samples, we conducted temperature ablation experiments on each sample, as presented in Table~\ref{tab:energykd_ablations}. Adjusting the temperature for both energy types yielded superior results compared to modifying the temperature for only one energy type, whether low-energy or high-energy.

We further explored the application of more varied temperatures across the entire dataset. To achieve this, we divided the entire CIFAR-100 dataset into 10 segments (i.e., $\textbf{x} \rightarrow [\textbf{x}_1,\textbf{x}_2,...,\textbf{x}_n, ...,\textbf{x}_{10}]$) based on their energy scores and applied different temperatures (i.e., $T_{1}, T_{2}, ..., T_{n}, ..., T_{10}$) to each segment. Specifically,

\begin{equation}
\mathbf{x}=\left\{\begin{array}{c}
\mathbf{x}_1 \rightarrow T_1=T_\mathrm{min} \\
\mathbf{x}_2 \rightarrow T_2 \\
\vdots \\
\mathbf{x}_{n} \rightarrow T_n \\
\vdots \\
\mathbf{x}_{10} \rightarrow T_{10}=T_\mathrm{max}
\end{array}\right..
\end{equation}
where $T_\text{min}$ and $T_\text{max}$ represent the minimum and maximum temperatures within the temperature range, respectively. We refer to this as \textit{Temperature Gradation} because we sequentially and gradually increase the temperature. (i.e., $T_{1}<T_{2}<...<T_{n}<...<T_{10}$)

Table~\ref{tab:energykd_ablations} demonstrates that the method with two different temperatures applied to both energy types achieves performance comparable to the \textit{Temperature Gradation}, which employs a broader range of temperature scaling. From these experiments, we can conclude that addressing certain and uncertain images is more meaningful than dealing with images in between. Additional details about each experiment can be found in Appendix.

\subsection{Sensitivity Analysis}
\label{subsec:sensitivity}

Table~\ref{table:sensitivity_r} demonstrates that Energy KD consistently outperforms KD across all values of the ratio $(r)$. Additionally, it is noteworthy that most models achieve optimal results at relatively low values of $r$. This suggests that prioritizing the processing of images with extreme energy values is more important than processing the majority of images.

\begin{table}[]
\centering
\resizebox{1.0\columnwidth}{!}{%
\begin{tabular}{cc|cccc}
\toprule
\multicolumn{2}{c|}{Teacher}                            & WRN-40-2       & ResNet32x4     & ResNet32x4     & VGG13          \\
\multicolumn{2}{c|}{Student}                            & WRN-16-2       & ResNet8x4      & ShuffleNetV2   & MobileNetV2    \\ \midrule
\multicolumn{2}{c|}{KD}                                 & 74.92          & 73.33          & 74.45          & 67.37          \\ \midrule
\multicolumn{1}{c|}{\multirow{5}{*}{Energy KD}} & $r=0.1$ & 75.24          & \textbf{74.62} & 75.32          & \textbf{68.85} \\
\multicolumn{1}{c|}{}                           & $r=0.2$     & \textbf{75.45} & 74.34          & \textbf{75.87} & 68.65          \\
\multicolumn{1}{c|}{}                           & $r=0.3$     & 75.29          & 74.60          & 75.76          & 68.75          \\
\multicolumn{1}{c|}{}                           & $r=0.4$     & 75.37          & 74.16          & 75.38          & 68.81          \\
\multicolumn{1}{c|}{}                           & $r=0.5$     & 75.06          & 74.28          & 75.65          & 68.70          \\ \bottomrule
\end{tabular}}
\caption{Sensitivity analysis on the values of the percentage $(r)$}
\label{table:sensitivity_r}
\end{table}

\subsection{Contribution of HE-DA}
\label{subsec:he-da_contribution}

\begin{table*}[t]
\centering
\resizebox{0.7\textwidth}{!}{%
\begin{tabular}{ccccccc}
\toprule
Teacher                            & WRN-40-2                              & ResNet50                             & ResNet32x4                           & VGG13                                & VGG13                                & ResNet32x4                           \\
Acc.                               & 75.61                                 & 72.34                                 & 79.42                                 & 74.64                                 & 74.64                                 & 79.42                                 \\
Student                            & WRN-16-2                              & ResNet20                             & ResNet8x4                            & VGG8                                 & MobileNetV2                           & ShuffleNetV2                          \\
Acc.                               & 73.26                                 & 69.06                                 & 72.50                                 & 70.36                                 & 64.60                                 & 71.82                                 \\ 
\midrule
KD*                                 & 74.92                                 & 70.66                                 & 73.33                                 & 72.98                                 & 67.37                                 & 74.45                                 \\
w/ CutMix** (100\%)                          & 75.34                                 & 70.77                                 & 74.91                           & 74.16                                 & \underline{68.79}                                 & 76.61                                 
\\
w/ HE-DA (20\%)  & 75.27                                 & 70.90                                 & 74.84                                 & 74.04                                 & 68.06                                 & 76.57                               \\
w/ HE-DA (30\%)  & 75.54                                 & 71.15                                 & 74.85                                 & 74.17                                 & 68.62                                 & 76.87                               \\
w/ HE-DA (40\%)  & \underline{75.72}                     & \textbf{71.43}                        & \underline{75.13}                     & \underline{74.42}                     & 68.69                     & \underline{77.16}                   \\
w/ HE-DA (50\%)  & \textbf{75.95}                        & \underline{71.26}                     & \textbf{75.22}                        & \textbf{74.54}                        & \textbf{69.13}                        & \textbf{77.32}                        \\
$\Delta$\textbf{*}                              & {   \textbf{+1.03}} & {   \textbf{+0.77}} & {   \textbf{+1.89}} & {   \textbf{+1.56}} & {   \textbf{+1.76}} & {   \textbf{+2.87}} \\  
$\Delta$\textbf{**}                              & {   \textbf{+0.61}} & {   \textbf{+0.66}} & {   \textbf{+0.31}} & {   \textbf{+0.38}} & {   \textbf{+0.34}} & {   \textbf{+0.71}} \\ 
\midrule
DKD*                                 & \underline{76.24}                                 & \underline{71.97}                                 & 76.32                                 & 74.68                                 & 69.71                                 & 77.07                                 \\
w/ CutMix** (100\%)                          & 75.72                                 & 71.59                                 & \underline{76.86}                           & \underline{75.14}                                 & \underline{70.81}                                 & \underline{78.81}\\
w/ HE-DA ($r\%$)  & \textbf{76.40}                        & \textbf{72.21}                     & \textbf{77.23}                        & \textbf{75.55}                        & \textbf{71.28}                        & \textbf{78.93}                        \\
$\Delta$\textbf{*}                              & {   \textbf{+0.16}} & {   \textbf{+0.24}} & {   \textbf{+0.91}} & {   \textbf{+0.87}} & {   \textbf{+1.57}} & {   \textbf{+1.86}}
\\
$\Delta$\textbf{**}                              & {   \textbf{+0.68}} & {   \textbf{+0.62}} & {   \textbf{+0.37}} & {   \textbf{+0.41}} & {   \textbf{+0.47}} & {   \textbf{+0.12}}
\\
\bottomrule
\end{tabular}%
}
\caption{Performance evaluated when applying High Energy-based Data Augmentation (HE-DA) to the CIFAR-100 test sets. The best results are highlighted in \textbf{bold} and the second best \underline{underlined}. 
$\Delta$\textbf{*} denotes the performance difference between the best result among various rates ($r\%$) of our method and the result without augmentation, while $\Delta$\textbf{**} denotes the performance difference between the best result and full data augmentation (100\%).}
\label{table:heda_cifar}
\end{table*}

Table~\ref{table:heda_cifar} showcases the outstanding performance of the High-Energy-based Data Augmentation (HE-DA) method on the CIFAR-100 dataset. Performance is evaluated by applying HE-DA to vanilla KD (refer to the upper table) and DKD (refer to the lower table), a state-of-the-art logit-based method. Results for augmenting the entire dataset (i.e., 100\%) are obtained from previous papers~\cite{augmentation, r2kd}.

In the case of vanilla KD, we achieve comparable performance to applying data augmentation to the entire dataset (i.e., 100\%), despite applying HE-DA to only 20\% of the data (i.e., $r=0.2$) for most models. The optimal performance of our method is reached when HE-DA is applied to 40-50\% of the data, resulting in a performance improvement of up to 2.87 over vanilla KD and up to 0.71 over that of data augmentation on the full dataset. Concerning DKD, our method attains a performance improvement of up to 1.86 over the baseline DKD and a performance improvement of up to 0.68 over that of data augmentation on the full dataset. Notably, when applying basic data augmentation methods (i.e., augmentation on the entire dataset) to DKD, some models perform worse than without augmentation. In contrast, our method consistently achieves performance improvements across all models.

\begin{table*}[t]
\centering
\resizebox{0.7\textwidth}{!}{%
\begin{tabular}{ccccccc}
\toprule
Teacher                            & WRN-40-2                              & ResNet56                             & ResNet32x4                           & VGG13                                & VGG13                                & ResNet32x4                           \\
Acc.                               & 61.28                                 & 58.37                                 & 64.41                                 & 62.59                                 & 62.59                                 & 64.41                                 \\
Student                            & WRN-16-2                              & ResNet20                             & ResNet8x4                            & VGG8                                 & MobileNetV2                           & ShuffleNetV2                          \\
Acc.                               & 58.23                                 & 52.53                                 & 55.41                                 & 56.67                                 & 58.20                                 & 62.07                                 \\ 
\midrule
KD*                                 & 58.65                                 & 53.58                                 & 55.67                                 & 61.48                                 & 59.28                                 & 66.34                                 \\
w/ CutMix** (100\%)                          & 59.06                                 & 53.77                                 & 56.41                           & 62.17                                 & 60.48                                 & 67.01                                 
\\
w/ HE-DA (20\%)  & 59.16                                 & \underline{54.09}                                 & 56.70                                 & 61.87                                 & 60.16                                 & 67.27                               \\
w/ HE-DA (30\%)  & 59.36                                 & 53.59                                 & 56.57                                 & \underline{62.36}                                 & \underline{60.63}                                 & \underline{67.45}                               \\
w/ HE-DA (40\%)  & \underline{59.54}                     & \textbf{54.52}                        & \underline{57.02}                     & 62.24                     & 60.59                     & 67.25                   \\
w/ HE-DA (50\%)  & \textbf{59.69}                        & 53.99                     & \textbf{57.13}                        & \textbf{62.51}                        & \textbf{60.85}                        & \textbf{67.64}                        \\
$\Delta$\textbf{*}                             & {   \textbf{+1.04}} & {   \textbf{+0.94}} & {   \textbf{+1.46}} & {   \textbf{+1.03}} & {   \textbf{+1.57}} & {   \textbf{+1.30}} \\ 
$\Delta$\textbf{**}                             & {   \textbf{+0.63}} & {   \textbf{+0.75}} & {   \textbf{+0.72}} & {   \textbf{+0.34}} & {   \textbf{+0.37}} & {   \textbf{+0.63}} \\  
\midrule
DKD*                                 & 59.66                                 & 54.39                                 & 58.57                                 & 63.12                                 & 61.70                                 & 67.37                                 \\
w/ CutMix** (100\%)                          & 59.92                                 & 54.01                                 & 59.23                           & 63.12                                 & 62.73                                 & 67.97\\
w/ HE-DA ($r\%$)  & \textbf{60.43}                        & \textbf{55.28}                     & \textbf{59.58}                        & \textbf{63.78}                        & \textbf{63.03}                        & \textbf{68.25}                        \\
$\Delta$\textbf{*}                              & {   \textbf{+0.77}} & {   \textbf{+0.89}} & {   \textbf{+1.01}} & {   \textbf{+0.66}} & {   \textbf{+1.33}} & {   \textbf{+0.88}}
\\
$\Delta$\textbf{**}                              & {   \textbf{+0.51}} & {   \textbf{+1.27}} & {   \textbf{+0.35}} & {   \textbf{+0.66}} & {   \textbf{+0.30}} & {   \textbf{+0.28}}
\\
\bottomrule
\end{tabular}%
}
\caption{Performance evaluated when applying High Energy-based Data Augmentation (HE-DA) to the TinyImageNet test sets. The best results are highlighted in \textbf{bold} and the second best \underline{underlined}. 
$\Delta$\textbf{*} denotes the performance difference between the best result among various rates ($r\%$) of our method and the result without augmentation, while $\Delta$\textbf{**} denotes the performance difference between the best result and full data augmentation (100\%).}
\label{table:heda_tiny}
\end{table*}

We extended our experiments to a more challenging dataset, TinyImagenet, to evaluate the performance of HE-DA, and the results are presented in Table~\ref{table:heda_tiny}. These results for TinyImagenet closely mirror those obtained for the CIFAR-100 dataset, showcasing excellent performance.

For vanilla KD (refer to the upper table), our method outperforms 100\% data augmentation despite applying only 20\%, with our best performance demonstrating an improvement of up to 1.57 over vanilla KD and up to 0.75 over 100\% data augmentation. Moving to DKD (refer to the lower table), our method achieves a performance improvement of up to 1.33 over basic DKD and up to 1.27 over 100\% data augmentation. Our method consistently delivers excellent performance across all models.

\begin{figure}[t]
	\centering
	\subfigure[VGG13-MV2\label{fig:vggmv}]
	{\includegraphics[width=0.45\columnwidth]{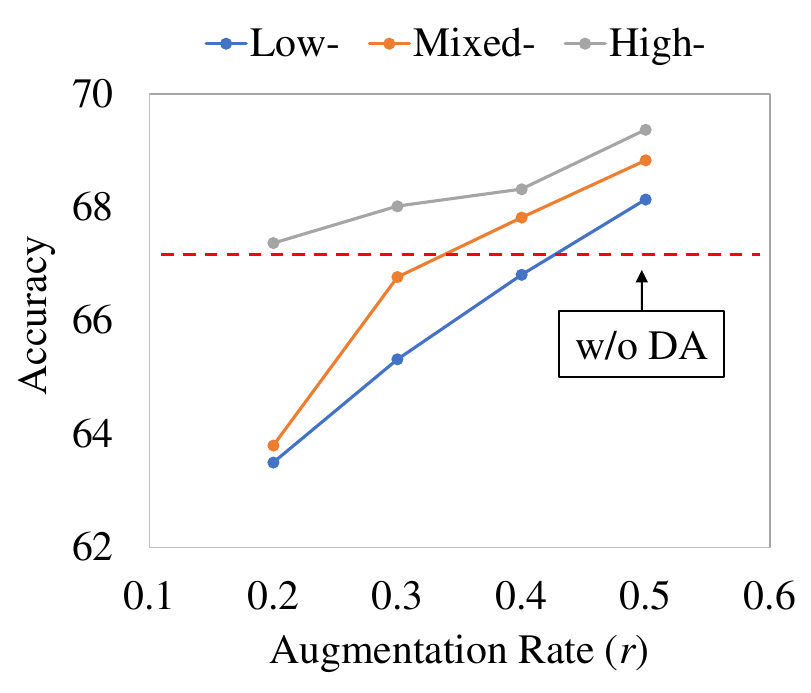}}
	\subfigure[Res32x4-SV2\label{fig:ressh2}]
	{\includegraphics[width=0.45\columnwidth]{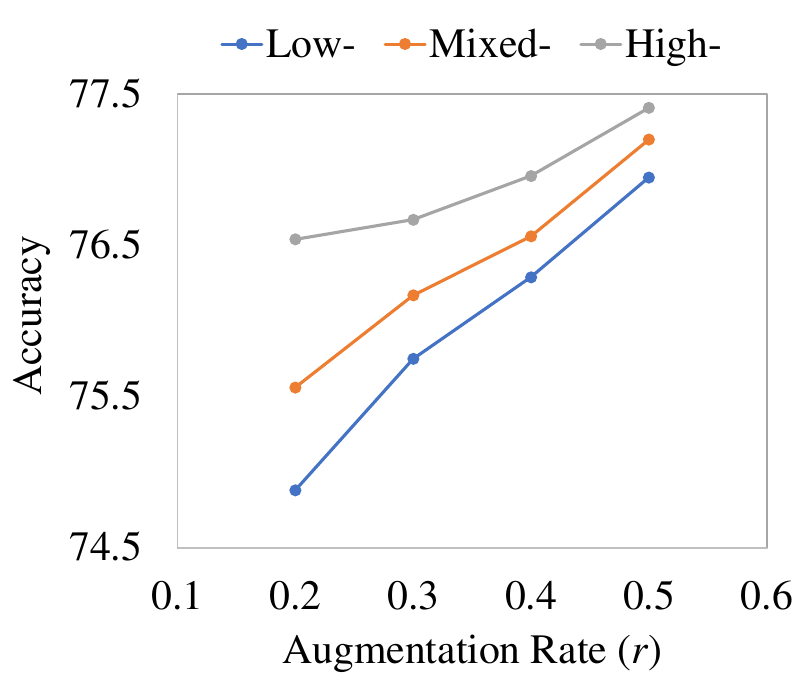}}
	\caption{Performance variations according to the sample types: low, high, and mixed energy. (a): VGG13/MobileNetV2, (b): ResNet32x4/ShuffleNetV2}
	\label{fig:ablation_same}
\end{figure}

Figure~\ref{fig:ablation_same} illustrates the performance variations based on sample types (i.e., low-energy, high-energy, and mixed-energy samples) for two different teacher-student pairs (i.e., VGG13-MobileNetV2 and ResNet32x4-ShuffleNetV2). These experiments clearly demonstrate that exclusively utilizing high-energy samples (marked as the gray line in the graph) results in higher performance compared to using low-energy samples (marked as the blue line) or a mix of samples, including half low-energy and half high-energy samples (marked as the orange line), for all augmentation rates $\left(r=0.2 \sim 0.5\right)$, which indicates the amount of additionally augmented samples for each type.

It is worth noting that when mixed samples of low-energy and high-energy data are employed (i.e., 50:50), the performance falls between the gray line (i.e., accuracy of high-energy samples) and the blue line (i.e., accuracy of low-energy samples). Using only high-energy data yields superior results, while using only low-energy data leads to lower performance, suggesting that reducing the additional low-energy data by augmentation positively affects performance. The reason behind this could be the learning model’s proficiency in understanding low-energy samples. Additional augmentation for low-energy samples, which already includes certain information for correct classification, might lead to confusion, hindering the learning process of the student model. Thus, for optimal performance, it is reasonable to decrease the quantity of augmented data for low-energy samples and rely solely on augmented data for high-energy samples.

Furthermore, it is worth noting that the accuracy of high-energy results remains relatively stable, regardless of the variation in augmentation rate, which is hyperparameters that determine the amount of augmented data. In other words, when dealing with high-energy data, results from augmenting 10\% to 20\% of the dataset show no significant difference compared to those from augmenting 40\% to 50\%. However, for low-energy data, augmenting 10\% to 20\% may result in lower performance compared to even results with no augmentation at all (marked as the red dash line), suggesting that the accuracy from using high-energy data is not significantly affected by changes in the augmentation rate. This characteristic can be particularly valuable in scenarios where computational resources are severely limited because using only a small amount of additional high-energy data is enough to achieve better results than previous KD methods.

\subsection{tSNE and Correlation}
\label{subsec:visualization}

\begin{figure}[]
	\centering
	{\includegraphics[width=0.8\columnwidth]{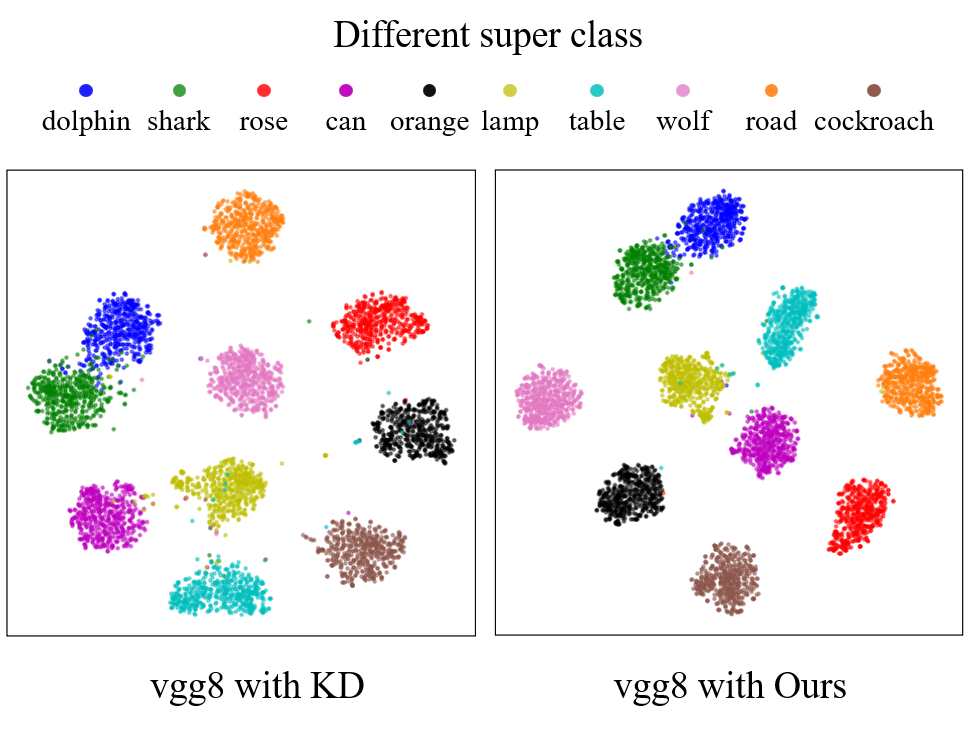}}
        {\includegraphics[width=0.8\columnwidth]{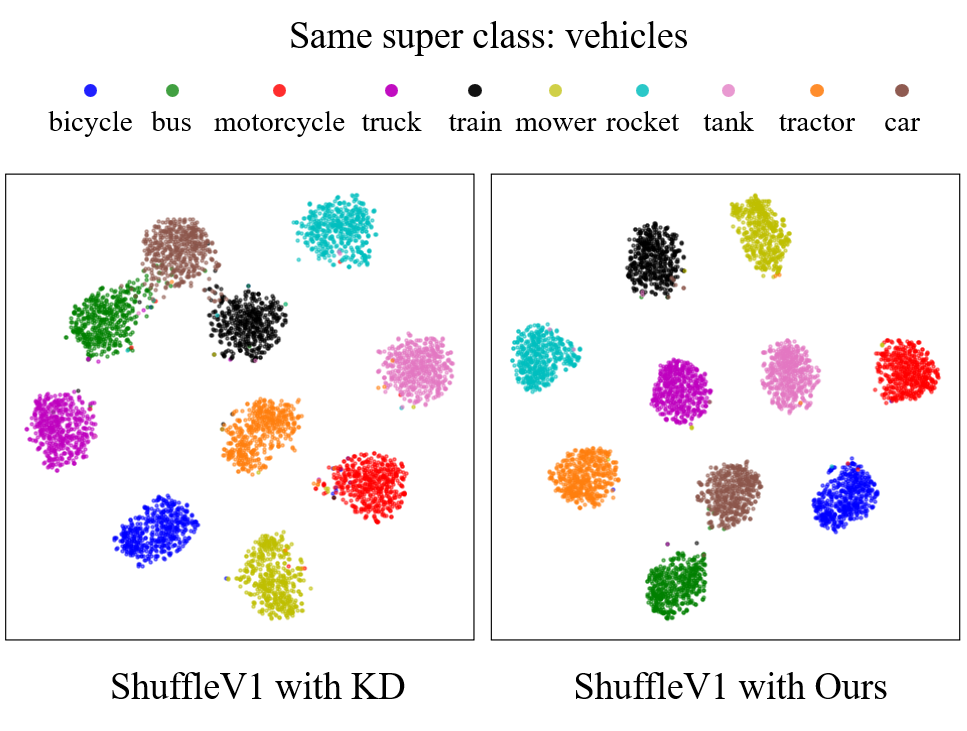}}
        {\includegraphics[width=0.8\columnwidth]{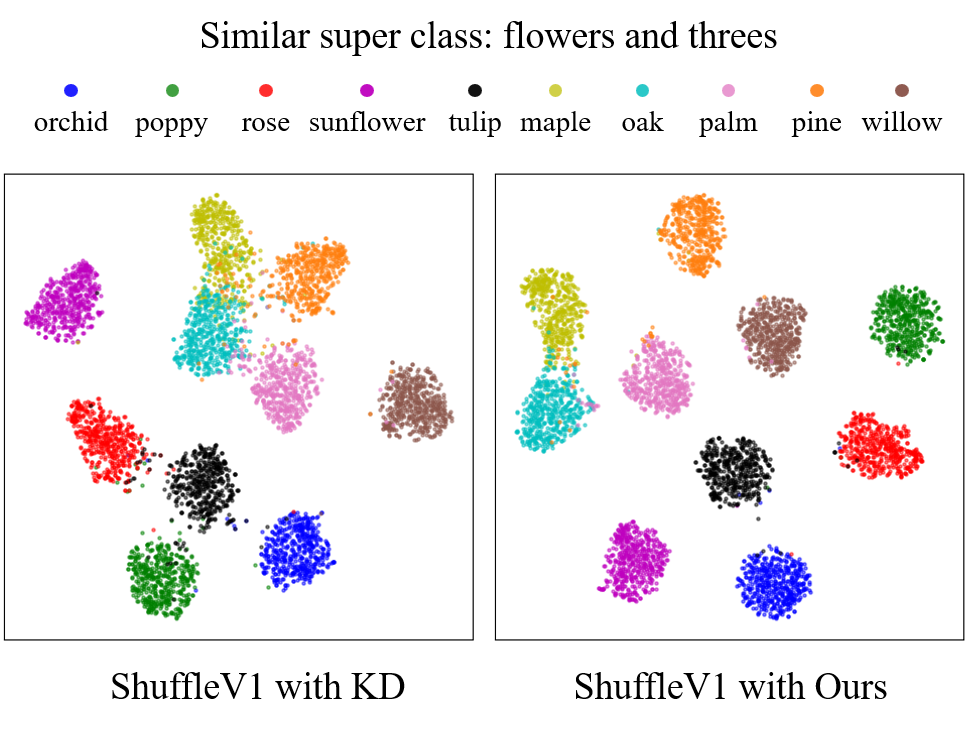}}
	\caption{Feature representations from the penultimate layer of student networks on the some classes of CIFAR-100 dataset. Upper: Different super class, Middle: Same super class (vehicles), Lower: Similar super class (flowers and threes).}
	\label{fig:tsne}
\end{figure}

\begin{figure}[]
	\centering
	\subfigure[KD]
	{\includegraphics[width=0.45\columnwidth]{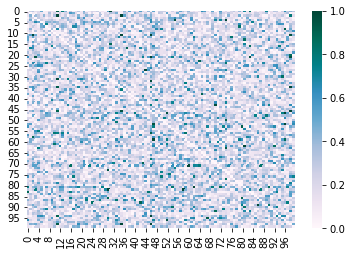}}
	\subfigure[Energy KD]
	{\includegraphics[width=0.45\columnwidth]{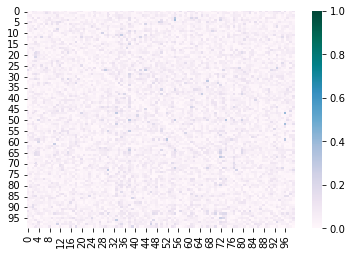}}
	\caption{Correlation disparities between the logits of the student and teacher. Energy KD shows smaller disparities than KD.}
	\label{fig:correlation}
\end{figure}

Figure~\ref{fig:tsne} displays tSNE results for various classes of CIFAR-100. The upper figure illustrates clustering for different super classes, showing little similarity between classes. The middle figure showcases clustering for the same super class (vehicles) containing similar classes, and the lower figure displays clustering for similar super classes (flowers and trees). In all figures, we observe that the representations produced by our method are closer for the same class and exhibit less overlap from other classes. Therefore, our method demonstrates better clustering ability compared to KD, enhancing the discriminability of deep features.

The essence of knowledge distillation lies in how closely the predictions of the student model align with those of the teacher model, given the information provided by the teacher. Figure~\ref{fig:correlation} visually illustrates this concept by comparing the correlation matrices of the student and teacher logits. Darker colors represent larger differences between the matrices, while lighter colors indicate smaller differences. In other words, the lighter the color, the better the student model mimics the teacher model and produces similar results, demonstrating its capability to yield superior outcomes. In contrast to previous KD, the application of our Energy KD induces the student to generate logits that are more similar to the teacher, thereby ensuring outstanding student performance.

\subsection{Computational Costs}
\label{subsec:computational costs}

Table~\ref{costs} presents the computational cost in relation to the percentage of augmentation applied, specifically referring to the learning time per epoch on the CIFAR-100 datasets. The table shows that applying augmentation to the entire dataset results in a 33.26\% increase in computational cost. When applying augmentation to 40-50\% of the dataset (which produces the peak performance of our method), we notice a more modest increase in computational expenses, ranging from 8.78\% to 14.17\%. These results demonstrate that our approach excels not only in terms of performance but also in terms of efficiency. Details regarding the computing infrastructure used for this experiment are introduced in Appendix.

\begin{table}[]
\centering
\resizebox{0.85\columnwidth}{!}{%
\begin{tabular}{c|cccccc}
\toprule
$r$       & 0.1  & 0.2   & 0.3   & 0.4   & 0.5    & 1.0    \\ \midrule
\% $\uparrow$ & 0.0\% & 3.94\% & 5.52\% & 8.78\% & 14.17\% & 33.26\% \\ \bottomrule
\end{tabular}%
}
\caption{Computational costs are measured according to the rate at which data augmentation is applied. The percentage rise is computed based on the value of $r=0.1$.}
\label{costs}
\end{table}

\subsection{Long-tailed Dataset}
\label{subsec:long-tailed}
Table~\ref{cifar100LT_energykd} presents the experimental results using the CIFAR-100 long-tail (LT) dataset. CIFAR-100-LT and CIFAR-100 are used for training and testing, respectively. The experiments involved four different architecture pairs. In the case of the long-tail type, exponential decay is applied, and an imbalance factor of 0.5 is used. More detailed experimental parameters are provided in Appendix. The table illustrates that our method outperforms state-of-the-art DKD and ReviewKD methods. These results highlight the effectiveness of our approach, even when dealing with challenging datasets, as observed in experiments on ImageNet datasets with significant differences in the number of samples among classes.

\begin{table}[h]
\centering
\resizebox{1.0\columnwidth}{!}{%
\begin{tabular}{ccccccc}
\toprule
Teacher                                & ResNet32x4                           & VGG13                                & VGG13                                & ResNet32x4                           
                             \\
Student                              & ResNet8x4                            & VGG8                                 & MobileNetV2                           & ShuffleNetV2                          
                            \\ \midrule
KD                                    & 72.51                                 & {{71.59}}                           & {{65.19} }                           & 72.93                                 \\
DKD                                      & 73.60                                  & 73.25                                 & 66.73                                 & 74.09                                 \\
ReviewKD                                         & {{73.42}}                           & 73.02                                 & 66.36                                 & {{74.11}}                           \\                                
Energy KD                          & \textbf{73.97}                        & \textbf{73.73}                        & \textbf{67.08}                        & \textbf{74.61}                        \\
\bottomrule
\end{tabular}%
}
\caption{Top-1 accuracy (\%) on the CIFAR-100-LT datasets, employing both identical and distinct architectures for teacher and student models.}
\label{cifar100LT_energykd}
\end{table}

\section{Conclusions}
\label{sec:conclusions}
In this paper, we introduce a novel perspective by incorporating the energy score of a sample, a factor traditionally overlooked. Our approach classifies datasets into low-energy and high-energy samples based on their energy scores, applying higher temperatures to low-energy samples and lower temperatures to high-energy samples. In comparison to both logit-based and feature-based methods, our EnergyKD consistently outperforms on various datasets. Notably, on challenging datasets such as CIFAR-100-LT and ImageNet, EnergyKD demonstrates significant performance gains, establishing its effectiveness in real-world scenarios. Furthermore, when coupled with High Energy-based Data Augmentation (HE-DA), it not only enhances performance but also maintains computational efficiency. We anticipate that our framework, offering a new perspective by considering the energy score of samples in both knowledge distillation and data augmentation, will pave the way for prosperous future research in model compression. However, this paper focuses solely on image classification. Extending our approach to various computer vision tasks, such as object detection and semantic segmentation, is part of our future work. For semantic segmentation, which involves pixel-by-pixel classification, devising an energy score function for each pixel will enable us to develop a segmentation-specific method.

  \bibliographystyle{elsarticle-num} 
  \bibliography{hakk}

\begin{thebibliography}{10}
\expandafter\ifx\csname url\endcsname\relax
  \def\url#1{\texttt{#1}}\fi
\expandafter\ifx\csname urlprefix\endcsname\relax\def\urlprefix{URL }\fi
\expandafter\ifx\csname href\endcsname\relax
  \def\href#1#2{#2} \def\path#1{#1}\fi

\bibitem{class1}
K.~He, X.~Zhang, S.~Ren, J.~Sun, Deep residual learning for image recognition, in: Proceedings of the IEEE conference on computer vision and pattern recognition, 2016, pp. 770--778.

\bibitem{class2}
N.~Ma, X.~Zhang, H.-T. Zheng, J.~Sun, Shufflenet v2: Practical guidelines for efficient cnn architecture design, in: Proceedings of the European conference on computer vision (ECCV), 2018, pp. 116--131.

\bibitem{obj1}
S.~Ren, K.~He, R.~Girshick, J.~Sun, Faster r-cnn: Towards real-time object detection with region proposal networks, Advances in neural information processing systems 28 (2015).

\bibitem{obj2}
K.~He, G.~Gkioxari, P.~Doll{\'a}r, R.~Girshick, Mask r-cnn, in: Proceedings of the IEEE international conference on computer vision, 2017, pp. 2961--2969.

\bibitem{seg1}
H.~Zhao, J.~Shi, X.~Qi, X.~Wang, J.~Jia, Pyramid scene parsing network, in: Proceedings of the IEEE conference on computer vision and pattern recognition, 2017, pp. 2881--2890.

\bibitem{seg2}
J.~Long, E.~Shelhamer, T.~Darrell, Fully convolutional networks for semantic segmentation, in: Proceedings of the IEEE conference on computer vision and pattern recognition, 2015, pp. 3431--3440.

\bibitem{pruning}
J.~Liu, B.~Zhuang, Z.~Zhuang, Y.~Guo, J.~Huang, J.~Zhu, M.~Tan, Discrimination-aware network pruning for deep model compression, IEEE Transactions on Pattern Analysis and Machine Intelligence 44~(8) (2021) 4035--4051.

\bibitem{quantization}
Y.~Zhou, S.-M. Moosavi-Dezfooli, N.-M. Cheung, P.~Frossard, Adaptive quantization for deep neural network, in: Proceedings of the AAAI Conference on Artificial Intelligence, Vol.~32, 2018.

\bibitem{kd_survey}
J.~Gou, B.~Yu, S.~J. Maybank, D.~Tao, Knowledge distillation: A survey, International Journal of Computer Vision 129 (2021) 1789--1819.

\bibitem{hinton}
G.~Hinton, O.~Vinyals, J.~Dean, Distilling the knowledge in a neural network, arXiv preprint arXiv:1503.02531 (2015).

\bibitem{dkd}
B.~Zhao, Q.~Cui, R.~Song, Y.~Qiu, J.~Liang, Decoupled knowledge distillation, in: Proceedings of the IEEE/CVF Conference on computer vision and pattern recognition, 2022, pp. 11953--11962.

\bibitem{review}
P.~Chen, S.~Liu, H.~Zhao, J.~Jia, Distilling knowledge via knowledge review, in: Proceedings of the IEEE/CVF Conference on Computer Vision and Pattern Recognition, 2021, pp. 5008--5017.

\bibitem{cifar}
A.~Krizhevsky, G.~Hinton, et~al., Learning multiple layers of features from tiny images (2009).

\bibitem{imagenet}
O.~Russakovsky, J.~Deng, H.~Su, J.~Krause, S.~Satheesh, S.~Ma, Z.~Huang, A.~Karpathy, A.~Khosla, M.~Bernstein, et~al., Imagenet large scale visual recognition challenge, International journal of computer vision 115 (2015) 211--252.

\bibitem{dml}
Y.~Zhang, T.~Xiang, T.~M. Hospedales, H.~Lu, Deep mutual learning, in: CVPR, 2018.

\bibitem{takd}
S.~I. Mirzadeh, M.~Farajtabar, A.~Li, N.~Levine, A.~Matsukawa, H.~Ghasemzadeh, Improved knowledge distillation via teacher assistant, in: AAAI, 2020.

\bibitem{jin2023multi}
Y.~Jin, J.~Wang, D.~Lin, Multi-level logit distillation, in: Proceedings of the IEEE/CVF Conference on Computer Vision and Pattern Recognition, 2023, pp. 24276--24285.

\bibitem{fitnet}
A.~Romero, N.~Ballas, S.~E. Kahou, A.~Chassang, C.~Gatta, Y.~Bengio, Fitnets: Hints for thin deep nets, arXiv preprint arXiv:1412.6550 (2014).

\bibitem{pkt}
N.~Passalis, M.~Tzelepi, A.~Tefas, Probabilistic knowledge transfer for lightweight deep representation learning, IEEE Transactions on Neural Networks and Learning Systems 32~(5) (2020) 2030--2039.

\bibitem{rkd}
W.~Park, D.~Kim, Y.~Lu, M.~Cho, Relational knowledge distillation, in: Proceedings of the IEEE/CVF Conference on Computer Vision and Pattern Recognition, 2019, pp. 3967--3976.

\bibitem{crd}
Y.~Tian, D.~Krishnan, P.~Isola, Contrastive representation distillation, arXiv preprint arXiv:1910.10699 (2019).

\bibitem{at}
S.~Zagoruyko, N.~Komodakis, Paying more attention to attention: Improving the performance of convolutional neural networks via attention transfer, arXiv preprint arXiv:1612.03928 (2016).

\bibitem{vid}
S.~Ahn, S.~X. Hu, A.~Damianou, N.~D. Lawrence, Z.~Dai, Variational information distillation for knowledge transfer, in: Proceedings of the IEEE/CVF Conference on Computer Vision and Pattern Recognition, 2019, pp. 9163--9171.

\bibitem{ofd}
B.~Heo, J.~Kim, S.~Yun, H.~Park, N.~Kwak, J.~Y. Choi, A comprehensive overhaul of feature distillation, in: Proceedings of the IEEE International Conference on Computer Vision, 2019, pp. 1921--1930.

\bibitem{fcfd}
D.~Liu, M.~Kan, S.~Shan, X.~CHEN, \href{https://openreview.net/forum?id=pgHNOcxEdRI}{Function-consistent feature distillation}, in: The Eleventh International Conference on Learning Representations (ICLR), 2023.
\newline\urlprefix\url{https://openreview.net/forum?id=pgHNOcxEdRI}

\bibitem{cat_kd}
Z.~Guo, H.~Yan, H.~Li, X.~Lin, Class attention transfer based knowledge distillation, in: Proceedings of the IEEE/CVF Conference on Computer Vision and Pattern Recognition, 2023, pp. 11868--11877.

\bibitem{lsh-tl}
J.~Li, Z.~Tang, K.~Chen, Z.~Cui, Knowledge distillation based on fitting ground-truth distribution of images, Applied Sciences 14~(8) (2024) 3284.

\bibitem{sakd}
Z.~Guo, P.~Zhang, P.~Liang, Sakd: Sparse attention knowledge distillation, Image and Vision Computing 146 (2024) 105020.

\bibitem{ackley1985learning}
D.~H. Ackley, G.~E. Hinton, T.~J. Sejnowski, A learning algorithm for boltzmann machines, Cognitive science 9~(1) (1985) 147--169.

\bibitem{salakhutdinov2010efficient}
R.~Salakhutdinov, H.~Larochelle, Efficient learning of deep boltzmann machines, in: Proceedings of the thirteenth international conference on artificial intelligence and statistics, JMLR Workshop and Conference Proceedings, 2010, pp. 693--700.

\bibitem{lecun2006tutorial}
Y.~LeCun, S.~Chopra, R.~Hadsell, M.~Ranzato, F.~Huang, A tutorial on energy-based learning, Predicting structured data 1~(0) (2006).

\bibitem{ranzato2006efficient}
M.~Ranzato, C.~Poultney, S.~Chopra, Y.~Cun, Efficient learning of sparse representations with an energy-based model, Advances in neural information processing systems 19 (2006).

\bibitem{ranzato2007unified}
M.~Ranzato, Y.-L. Boureau, S.~Chopra, Y.~LeCun, A unified energy-based framework for unsupervised learning, in: Artificial Intelligence and Statistics, PMLR, 2007, pp. 371--379.

\bibitem{zhao2016energy}
J.~Zhao, M.~Mathieu, Y.~LeCun, Energy-based generative adversarial network, arXiv preprint arXiv:1609.03126 (2016).

\bibitem{xie2018cooperative}
J.~Xie, Y.~Lu, R.~Gao, S.-C. Zhu, Y.~N. Wu, Cooperative training of descriptor and generator networks, IEEE transactions on pattern analysis and machine intelligence 42~(1) (2018) 27--45.

\bibitem{xie2018learning}
J.~Xie, Z.~Zheng, R.~Gao, W.~Wang, S.-C. Zhu, Y.~N. Wu, Learning descriptor networks for 3d shape synthesis and analysis, in: Proceedings of the IEEE conference on computer vision and pattern recognition, 2018, pp. 8629--8638.

\bibitem{xie2019learning}
J.~Xie, S.-C. Zhu, Y.~N. Wu, Learning energy-based spatial-temporal generative convnets for dynamic patterns, IEEE transactions on pattern analysis and machine intelligence 43~(2) (2019) 516--531.

\bibitem{liu2020energy}
W.~Liu, X.~Wang, J.~Owens, Y.~Li, Energy-based out-of-distribution detection, Advances in neural information processing systems 33 (2020) 21464--21475.

\bibitem{energyood}
W.~Liu, X.~Wang, J.~Owens, Y.~Li, Energy-based out-of-distribution detection, Advances in neural information processing systems 33 (2020) 21464--21475.

\bibitem{energy1}
W.~Grathwohl, K.-C. Wang, J.-H. Jacobsen, D.~Duvenaud, M.~Norouzi, K.~Swersky, Your classifier is secretly an energy based model and you should treat it like one, arXiv preprint arXiv:1912.03263 (2019).

\bibitem{dark1}
D.~Y. Park, M.-H. Cha, D.~Kim, B.~Han, et~al., Learning student-friendly teacher networks for knowledge distillation, Advances in neural information processing systems 34 (2021) 13292--13303.

\bibitem{dark2}
C.~Yuan, R.~Pan, Obtain dark knowledge via extended knowledge distillation, in: 2019 International Conference on Artificial Intelligence and Advanced Manufacturing (AIAM), IEEE, 2019, pp. 502--508.

\bibitem{ATS}
X.-C. Li, W.-S. Fan, S.~Song, Y.~Li, S.~Yunfeng, D.-C. Zhan, et~al., Asymmetric temperature scaling makes larger networks teach well again, Advances in Neural Information Processing Systems 35 (2022) 3830--3842.

\bibitem{cutmix}
S.~Yun, D.~Han, S.~J. Oh, S.~Chun, J.~Choe, Y.~Yoo, Cutmix: Regularization strategy to train strong classifiers with localizable features, in: Proceedings of the IEEE/CVF international conference on computer vision, 2019, pp. 6023--6032.

\bibitem{mixup}
H.~Zhang, M.~Cisse, Y.~N. Dauphin, D.~Lopez-Paz, mixup: Beyond empirical risk minimization, arXiv preprint arXiv:1710.09412 (2017).

\bibitem{resnet}
K.~He, X.~Zhang, S.~Ren, J.~Sun, Deep residual learning for image recognition, in: CVPR, 2016.

\bibitem{wideresnet}
S.~Zagoruyko, N.~Komodakis, Wide residual networks, in: BMVC, 2016.

\bibitem{vgg}
K.~Simonyan, A.~Zisserman, Very deep convolutional networks for large-scale image recognition, in: ICLR, 2015.

\bibitem{mobilenet}
M.~Sandler, A.~Howard, M.~Zhu, A.~Zhmoginov, L.-C. Chen, Mobilenet{V}2: {I}nverted residuals and linear bottlenecks, in: CVPR, 2018.

\bibitem{shufv1}
X.~Zhang, X.~Zhou, M.~Lin, J.~Sun, Shufflenet: An extremely efficient convolutional neural network for mobile devices, in: Proceedings of the IEEE conference on computer vision and pattern recognition, 2018, pp. 6848--6856.

\bibitem{shufv2}
N.~Ma, X.~Zhang, H.-T. Zheng, J.~Sun, Shufflenet v2: Practical guidelines for efficient cnn architecture design, in: Proceedings of the European conference on computer vision (ECCV), 2018, pp. 116--131.

\bibitem{augmentation}
H.~Wang, S.~Lohit, M.~N. Jones, Y.~Fu, What makes a" good" data augmentation in knowledge distillation-a statistical perspective, Advances in Neural Information Processing Systems 35 (2022) 13456--13469.

\bibitem{r2kd}
S.~Kim, G.~Ham, Y.~Cho, D.~Kim, Robustness-reinforced knowledge distillation with correlation distance and network pruning, arXiv preprint arXiv:2311.13934 (2023).

\end{thebibliography}






\end{document}